\documentclass[final,5p,times,twocolumn]{elsarticle}
\makeatletter 
\def\ps@pprintTitle{  
    \let\@oddhead\@empty  
    \let\@evenhead\@empty  
    \def\@oddfoot{\hfill\thepage}
    \def\@evenfoot{\thepage\hfill}} 
\makeatother
\usepackage{amssymb}
\usepackage{amsmath}
\usepackage{graphics}
\usepackage{graphicx}
\usepackage{booktabs}
\usepackage[autostyle]{csquotes}
\usepackage{url}
\usepackage{epsfig}
\usepackage{float}
\usepackage{soul}
\usepackage{caption}
\usepackage{subcaption}
\usepackage{xcolor, soul}
\sethlcolor{green}

\begin{document}

\begin{frontmatter}

\title{Onto4MAT: A Swarm Shepherding Ontology for Generalised Multi-Agent Teaming}

\author{Adam J. Hepworth}\corref{cor1}
\ead{a.j.hepworth@unsw.edu.au}
\author{Daniel P. Baxter}
\ead{daniel.baxter@student.adfa.edu.au}
\author{Hussein A. Abbass}
\ead{h.abbass@unsw.edu.au}

\cortext[cor1]{Corresponding author}

\address{School of Engineering and Information Technology, University of New South Wales, Canberra, ACT, 2600, Australia}

\begin{abstract}
    Research in multi-agent teaming has increased substantially over recent years, with knowledge-based systems to support teaming processes typically focused on delivering functional (communicative) solutions for a team to act meaningfully in response to direction. Enabling humans to effectively interact and team with a swarm of autonomous cognitive agents is an open research challenge in Human-Swarm Teaming research, partially due to the focus on developing the enabling architectures to support these systems. Typically, bi-directional transparency and shared semantic understanding between agents has not prioritised a designed mechanism in Human-Swarm Teaming, potentially limiting how a human and a swarm team can share understanding and information\textemdash data through concepts and contexts\textemdash to achieve a goal. To address this, we provide a formal knowledge representation design that enables the swarm Artificial Intelligence to reason about its environment and system, ultimately achieving a shared goal. We propose the Ontology for Generalised Multi-Agent Teaming, \textit{Onto4MAT}, to enable more effective teaming between humans and teams through the biologically-inspired approach of shepherding.
\end{abstract}

\end{frontmatter}

\section{Introduction}\label{sec:intro}

Human-autonomy teaming (HAT) is a term used to describe humans and intelligent, autonomous agents working interdependently toward a common goal~\cite{Chen2016, Waynne2018}. McNeese specifies HAT as at least one human working cooperatively with at least one autonomous agent~\cite{McNeese2018}. Studies of HAT scenarios show that teaming tendency can be so influential that humans align their responses with their autonomous teammates compared to independent performance~\cite{Panganiban2020}.

According to Endsley~\cite{Endsley2017}, multi-agent teaming (MAT) extends HAT to include multiple autonomous agents, both human and artificial, working independently of each other without the need for intervention. Further, agents in MAT extend the HAT agents behavioural corpus from support roles to focus on asynchronous problem solving~\cite{Flathmann2019}, which is complex for heterogeneous, multi-agent teams with many challenges to overcome. Several methods have been proposed in MAT literature to maximise the quality of agents' solutions through individual qualities to improve team performance~\cite{Andrejczuk2017, Baxter2021}. Correct implementation of these teammate qualities is essential to facilitate a robust teamwork process and, subsequently, taskwork. The principles of teamwork and taskwork have a deep foundation in the human-human teaming literature, where taskwork describes \textit{what} it is the team is doing, and teamwork describes \textit{how} they are doing it~\cite{Marks2001}. Marks goes on to define the teamwork process as \textcquote[pg. 2]{Marks2001}{members’ interdependent acts that convert inputs to outcomes through cognitive, verbal, and behavioural activities directed toward organising taskwork to achieve collective goals.}

Human-Swarm Teaming (HST) takes MAT one step further to teaming with swarms. Swarms are typically an explicitly or implicitly synchronised team of autonomous agents with the ability to self organise and generate collective global behaviours from decentralised local behaviours~\cite{Hussein:2018}. HST is an emerging field of research, burgeoning with an increasing prevalence of highly competent autonomous agents interacting with humans~\cite{Hepworth2021:HST3}.

The individuals within HST and the overall teaming arrangements in HST require guidance mechanisms. Humans may need to guide the swarm, and an external entity may need to guide the teaming arrangements among the humans and the swarm. Shepherding is a biologically-inspired swarm guidance and control technique~\cite{Abbass2020:ShepherdingUxVs, Long2020:Comprehensive}), whereby one or a few powerful cognitive agents guide a larger group of potentially less-powerful agents, similarly to that of a sheepdog guiding a flock of sheep~\cite{Hepworth2021:ARS}.

The context of shepherding for HST is vibrant with many categories of concepts. It is paramount to organise domain concepts for processing, where an ontology enables an AI shepherding agent to deal with more complex scenarios. Ontologies provide a systematic way to decompose a system with hierarchies of concepts, each with properties that describe meaning in the domain~\cite{golenkov2017ontology, Guarino2009:ontologybook}.

\begin{figure*}[!h]
    \centering
    \includegraphics[width=\textwidth]{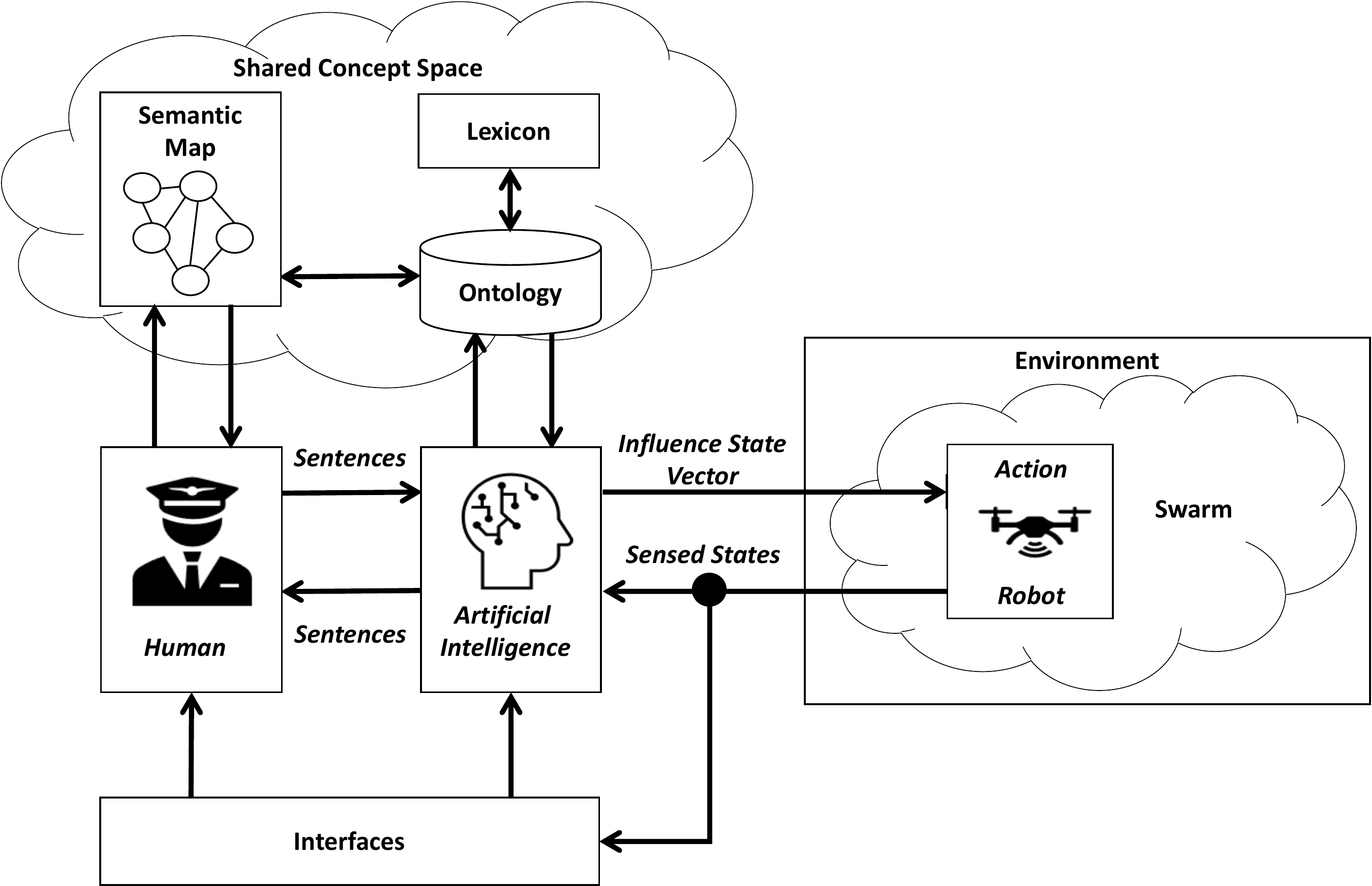}
    \caption{The conceptual flow of information as sentences exchanged between humans and AI agents. This bi-directional flow grounds meaning for both human and AI agents to generate shared understanding.}
    \label{fig:grounding}
\end{figure*}

For the cognitive agents in the system (be they human or artificial), an ontology refines knowledge structure~\cite{Methontology:1997}, reduces conceptual and lexical ambiguity~\cite{Oberle2009}, and provides a platform for knowledge sharing~\cite{GRUBER1995907}. For HST, an ontology can contribute to the formation of shared understanding among HST members, support the natural language functions required to exchange and negotiate meaning during bi-directional communications, offer a layer of transparency, and ultimately increase trust among members of the HST. Moreover, an ontology could guide the system as it transforms raw sensorial information to actions by assigning meaning to functions, guiding the system towards unexplored conceptual sub-spaces, and providing a means to assure the system by assuring the implementation of individual concepts. Ontologies bridge the semantic gap between a human and an artificial agent. As the agents exchange syntactical structures, the artificial agent relies on the ontology to connect concepts and ultimately to understand context, which offers a transparent form of shepherding as depicted in Figure~\ref{fig:grounding}.

This paper is the first attempt to design an ontology for MAT systematically. Previous attempts in shepherding and robotics focused purely on the guidance and control component of the system. However, bi-directional and multi-directional communications require the agents to have a shared ontology to ensure a mutual and shared understanding of the interaction. Our generic ontology for MAT (Onto4MAT) is instantiated specifically for shepherding as a use case to illustrate how to contextualise it for a particular swarm guidance task.

In the remainder of this paper, Section~\ref{sec:literature} reviews the contemporary literature on swarm-related ontologies and HST systems. Section~\ref{sec:design} introduces our proposed generic ontology architecture, describing the methodology to develop contexts and situations, followed by the presentation of Onto4MAT in Section~\ref{sec:onto4mat}. We describe an evaluation methodology to develop additional contexts in Section~\ref{sec:eval}, prior to concluding the paper Section~\ref{sec:conclusion}.

\section{Background Materials}\label{sec:literature}

\subsection{Ontological Approaches for Teaming}

Knowledge-based systems are widely used to support the design of Multi-Agent System (MAS) frameworks across a broad range of settings and domains~\cite{ZHA2002493}. Ontologies are established as a method to deliver understanding, particularly in settings of MAS, Human-Robot Interaction (HRI) and HAT, which are an \enquote{essential activity} as autonomous agents \enquote{should follow human requests}~\cite{ZHANG2007310}. Ontologies have been designed for various HRI and related purposes; however, few directly address HST and swarming application areas more broadly. Swarm Ontology~\cite{Abbass2020:ShepherdingUxVs} presents higher-level concepts, which link to the foundations of HST.

GAIA~\cite{GAIA} was presented in 2000 as a general methodology applicable to a wide range of MAS situations. As a conceptual framework, it was designed to enable an analyst to go from requirements to design seamlessly, considering some level of MAS interaction. In 2002, ADELFE~\cite{ADELFE} aimed to present a methodology that would aid a designer to build an adaptive MAS based on AMAS theory. ADELFE considers cooperation between agents; however, its implementation as a software development methodology excluded humans from the MAS context. It was not until 2008 when MOBMAS~\cite{TRAN2008697} was presented, focusing design for MAS. MOBMAS lacks considerations for HST and HMT situations whereby humans are represented as agents, particularly for teaming situations.

Moving from the abstraction of MAS to physical embodiment of cognition in robots, Tenorth and Beetz~(2009)~\cite{KnowRob} present a \enquote{knowledge processing system specifically designed for autonomous robots}. Based on first-order predicate logic, the ontology is an action-centric representation. The KnowRob ontology abstracts individual robotic actions focusing on improving individual, not collective behaviours and coordination~\cite{olivares:2019}. The ORO project\cite{ORO:2010} designed a framework for common representation across autonomous robotics, emphasising the interaction between humans and robots. The framework has been the basis for interaction and communication approaches between humans and robots~\cite{olivares:2019}. RobotML~\cite{RobotML} was proposed in 2012 as a way to enable scientific knowledge transfer between different robotic communities. While not designed to promote transparency in HST or human-machine teaming (HMT) scenarios, it did provide a domain-specific language to assist robotic designers.

While the above literature focuses on MAS and robotic systems, little is published focusing on swarms and less for HST, which is our focus in this paper. SWARMS~\cite{SWARMs:2017} is presented as an ontology of ontologies to facilitate unambiguous information exchange. It is presented as a way for heterogeneous data to be shared transparently, although the machine-specific focus could present a substantial barrier to adoption beyond the particular application domain. The IEEE 1872 Standard Ontology for Autonomous Robotics (CORA)~\cite{IEEE1872-2015} provides an overarching ontology framework and associated methodology which extends CORA as a way to represent and reason for autonomous robotic systems~\cite{Olszewska2017}. The framework presents a standard that aims to serve as a way for humans and robots to share information; however, it is considered too complex for general applications~\cite{olivares:2019}.

The area of shepherding has witnessed recently work on ontologies. In particular, Abbass and Hunjet~(2021)~\cite{Abbass2020:ShepherdingUxVs} introduce an ontology to relate core concepts required by an individual or team of swarm control agents. A primary motivation is to promote transparency through a knowledge-driven approach to explanations, a common semantic language (interpretable) and assurance by a human. This allows the \enquote{human to learn from it, use it and to assure its trustworthiness}~\cite{Abbass2020:ShepherdingUxVs}.

Contemporary studies are broadly motivated to aid system designers to develop HRI and MAS ontologies, implementing architectures at a functional level; however, they are seldom explicitly designed to enable teaming\textemdash understanding \textemdash between human and machine agents~\cite{KnowRob, SWARMs:2017, ORO:2010, RobotML, IEEE1872-2015}. Higher-level concepts, such as those of mission and intent, are often insufficiently prioritised in the available literature. Furthermore, the grounding of concepts that generate shared understanding in group scenarios, such as tactics, formations, roles, and team responsibilities, are absent. In the next section, we critically assess the previous ontologies concerning their stated abilities to any explicit or inferred HST contexts.

\subsection{Critical Assessment of Ontological Approaches for Teaming}

GAIA~\cite{GAIA} is a framework that enables MAS interaction, but it does not extend to HST. ADELFE~\cite{ADELFE} deals with concepts for dynamic and evolving environments. Communication between agents is not designed to support bi-directional teaming, such as in HST settings. MOBMAS~\cite{TRAN2008697} considers the functional design and implementation of MAS humans may be represented as agents, although no teaming aspects and communication structures are provided for HST, by design. KnowRob~\cite{KnowRob} is the basis of a later approach for interaction between humans and robots~\cite{olivares:2019}.

KnowRob's shallow symbolic representations improve individual agent autonomy; however, they may not enable cooperation and coordination through shared understanding in MAS~\cite{SWARMs:2017}. The ontology prioritises the enhancement of a robotic system from the agent's perspective, which, while improving the agent's cognition, may not enhance the ability of a human to interact as part of HST. ORO~\cite{ORO:2010} discusses human-robot interaction with autonomous agents possessing advanced cognitive skills. The autonomous system is assumed to be in a supporting role (uni-directional) for a human and may not be generating the necessary functions required for teaming. This uni-directional communication limits its ability to obtain enhanced transparency among agents. RobotML~\cite{RobotML} is formulated in such a way that presents complex issues for implementation and direct reasoning. The stated purpose of RobotML is to enable scientific knowledge transfer at the functional implementation level, not to support HST outcomes.

SWARMS~\cite{SWARMs:2017} offers a way to share information between heterogeneous agents. The design is agnostic of human interaction, focusing on robotic MAS in a particular domain, potentially limiting agent transparency as part of HST. CORA~\cite{IEEE1872-2015} is designed with communication central to the aim of the ontology. CORA provides the functional implementations required to enable information sharing between agents, agnostic of the application domain. A shared conceptualisation of semantic information for understanding is given. The generality of the ontology presents application difficulties in more complex domains and situations, such as HST. While Abbass and Hunjet's~(2021)~\cite{Abbass2020:ShepherdingUxVs} ontology captures the core capabilities of swarm control agents for HST, such as individual and collective behavioural sets and tactics and higher-level strategies such as behavioural synchronisation, it is a specific ontology for shepherding that may not generalise to broader MAT scenarios.

A common theme from our review is that contemporary ontologies do not fully account for guidance and control in teaming situations, highlighting a research gap specifically for HST. Current ontologies typically focus on functional specifications. Previous work has identified transparency as a critical enabler to generate trust in HST settings~\cite{Hepworth2021:HST3, Abbass2021:Symbiomemesis}, requiring more than uni-directional information flow, necessitating bi-directional communication for the generation of shared understanding with information. However, agent capabilities are functionally related, which may not capture the requirements to enable transparent, bi-directional HST. There is little discussion in the literature of the design and implementation of an ontology to support HST, particularly in situations of humans dealing with cognitive agents for swarm control. We hypothesise that there is a requirement for an ontology to translate between biological and artificial agents to control a swarm of cognitive agents. The remainder of this paper focuses on achieving the aim of designing an ontology for MAT.

Future teaming scenarios will require artificial (information and physical) and biological (human and possibly animals) agents to seamlessly share information to achieve meaningful interactions. Such sharing of information improves situational awareness to facilitate team interactions, establish expectations and enable future interactions~\cite{chen2018}. In settings where teaming between biological and artificial agents is essential, we propose that a generalised knowledge architecture will be required to maintain situational understanding across different contexts and for different functions. An ontology captures the key concepts in this interaction and could describe the skills and traits within the context to generate a shared understanding of the semantics underlying a  domain~\cite{chen2018}. While information fusion defines the functional requirements for synthesising the information required by an agent to perform a task, the lower levels of the fusion architecture do not provide sufficient information on meaning to allow agents to develop a mutual understanding of context and situations.

Girardi and Leite (2008)~\cite{GIRARDI2008604} discuss ways to contextualise an agent's roles, responsibilities, interactions, activities, goals, knowledge and skill. They note that domain concepts and their relationships are represented in agent-oriented models. This approach has inspired our work to prioritise the contextual understanding formed between agents in teaming situations. This paper uses shepherding as the vehicle to design an ontology to support MAT. The Human-Swarm-Teaming Transparency and Trust (HST3) Architecture was introduced by Hepworth et al.~(2021)~\cite{Hepworth2021:HST3} to promote transparency as an enabler for effective HST. Hepworth et al.~(2021) assert that \enquote{maintaining a high Situational Awareness (SA) is an enabler for human decision making.} A relationship is established between the increasing cognitive capabilities of autonomous agents and the increased demand for effective and efficient bi-directional interaction. The HST3-Architecture consists of a three-tiers being a lower layer (agent knowledge), a middle layer (inference engine), and a top layer (communication tier). The top layer delivers interpretability, explainability and predictability, and knowledge and model base modules at the agent logic level. Each higher-level architecture module delivers a portfolio of capabilities such as communication and language services, swarm guidance and control functions, inference services, and context-awareness services. HST3 is a frame to share semantic information, leading to effective teaming outcomes between biological and artificial agents.
 
\section{Methodology}\label{sec:designMethod}

\subsection{Ontology Design and Evaluation}\label{SubSection:designAndEval}

The conceptualisation of a domain instantiates formally-presented knowledge as concepts, functions, objects, properties, and relationships. The representation of this knowledge forms everything that \enquote{exists} in the world for each agent in the domain~\cite{GRUBER1995907}. The elements components of this knowledge set differ between research fields, domains of application, philosophical stances, and ontology languages used~\cite{Noy2001:Ontology101}. In this paper, we align our knowledge components to be consistent with the artificial intelligence literature and describe an ontology, $\mathcal{O}$, as a five-tuple $\mathcal{O} = <C,R,a,I,A>$ consisting of: concepts ($C$), relationships ($R$), attributes ($a$), instances ($I$), and axioms ($A$). Domain concepts are described ontologically through classes, which are the main focus of most ontologies as they formalise the highest-level elements of the domain. Relationships are the links between the concepts to represent the structure of the ontology and provide meaning as a set of linked definitions~\cite{IDEF5}. Attributes are the properties and characteristics that assert an ontology's data to discriminate and reason in the environment. Instances, also known as individuals or objects, are the lowest-level components of an ontology and is ultimately the element ontologies set out to describe or classify~\cite{Reyes2019}. Axioms are given in the Semantic Web Rule (SWR) Language (SWRL), presented as constraints to limit possible \enquote{interpretations for the defined terms (p.3)}~\cite{IDEF5}.

The Methontology framework~\cite{Methontology:1997} was selected by the IEEE SA P1872.2 Autonomous Robotics (AuR) Ontology Working Group to extend the CORA ontology, chosen due to its mature ontology development methodology~\cite{IEEE1872-2015}. We selected it in conjunction with the IDEF5 ontology description capture methodology~\cite{IDEF5}. The two design methodologies, while independently published, share key concepts and insights that we adopt to inform our design methodology presented in Section~\ref{sec:design}.

Knowledge acquisition is an independent but coincident activity that is particularly important in the early stages of ontology development. The initial domain glossary can be informed by texts, figures, publications, and previously developed ontologies. The domain knowledge can be elicited through brainstorming sessions, informal and formal interviews with domain experts, and inspecting previous ontologies. Knowledge acquisition has a diminishing contribution as development continues into and beyond the conceptualisation stage. Conceptualisation is where we structure the previously acquired domain knowledge into a model that describes the problem and its solution. Here, we expand the initial glossary to include concepts, instances, and properties to structure a semi-formal specification using Intermediate Representations (IRs) in the form: (a) the full glossary of terms; (b) a concept classification tree; (c) a binary-relations diagram; (d) the concept dictionary; (e) a binary-relations table; (f) an instance-attribute table; (g) a logical-axioms table; (h) a formula table; (i) an attribute-classification tree; and, (j) part of an instance table~\cite{Lopez1999}. Throughout the first two stages in the design method, we identified a set of candidate terms and definitions that could be integrated from existing meta-ontologies. Of particular note, the Human-Swarm Teaming Guidance Ontology (HST-GO)~\cite{Baxter:HST-GO}, Swarm Ontology for Shepherding~\cite{Abbass2020:ShepherdingUxVs}, and SOSA: a lightweight ontology for sensors, observations, samples, and actuators~\cite{JANOWICZ20191} formed the core of existing ontologies that provided definitions of terms whose semantics and implementation aligns with the terms identified in our conceptualisation of the domain. Ontology implementation is the codification of the ontology using formal representations in Prolog, Ontolingua, C++, or XML~\cite{Lopez1999}, or more recently, Python~\cite{onto:py}. Many environments exist to support the implementation of ontologies that include additional tools to detect lexical and syntactic errors, reasoners for logical errors, library browsers and modifiers, and evaluations for incomplete and inconsistent knowledge. Individual evaluations analyse an ontology broadly; best practice literature details an evaluation plan consisting of a multi-approach verification and validation method for an in-depth analysis of ontology development during and between phases.

Ontology evaluation is a growing field with importance highlighted in several proposed evaluation methodologies ~\cite{Tartir2010:Evaluation, GomezPerez2001, Hartmann2004, Staab2004, Lovrencic2008}. Evaluation methodology selection depends on factors such as the degree of formality~\cite{Kabilan2006}, the domain of interest, or future extensions of the ontology~\cite{Lovrencic2008}. Some ontology development methodologies include evaluation following integration or implementation. Others, such as Methontology, propose evaluation to be systemic throughout the ontology development process. We adopt G\'omez-P\'erez\textquoteright s evaluation definition~\cite{GomezPerez2004} as \enquote{a technical judgement of the content of the ontology concerning a frame of reference during every phase and between phases of their life cycle.} Where a frame of reference may consist, among others, a requirements specification, workshop questions, or the real world, and will likely reflect the current life cycle evaluation stage—further, we dissect our evaluation to it\textquoteright s two constituents: verification and validation. We adopt G\'omez-P\'erez\textquoteright s definition of verification~\cite{GomezPerez2004} as \enquote{referring to building the ontology correctly, that is, ensuring that its definitions (written in informal or formal language) implement correctly the ontology requirements and competency questions, or function correctly in the real world,} and validation~\cite{GomezPerez2004} as \enquote{referring to whether the ontology definitions model the real world for which the ontology was created. The goal is to prove that the world model (if it exists and is known) is compliant with the world modelled formally.}

Ontology evaluation methods can utilise a combination of several criteria. Many tools exist to help assess ontologies against these criteria. We select two independent evaluation methods: OntoClean, and OOPS!, as discussed in \ref{SubSection:designAndEval} to evaluate the ontology from different perspectives\textemdash from philosophical logics and ontology modelling to taxonomic representation. Verifying taxonomic knowledge in ontologies helps identify errors in definitions and axioms. The outcome is two-fold, firstly, to determine what the ontology defines correctly, what it defines incorrectly, and what it fails to define. Secondly, what can be inferred, what is incorrectly inferred, and what it fails to infer.

OntOlogy Pitfall Scanner (OOPS!) is a semi-autonomous diagnosis tool that verifies OWL ontologies from a different perspective. It is an online tool that operates independently of ontology development platforms to detect pitfalls in the ontology description logic that may lead to modelling errors. The tool has a catalogue of 40 pitfalls classified according to their structural, functional, and useability-profiling dimensions. In the event OOPS! detects a pitfall, a level of importance (critical, important, minor) is determined based on the impact of the pitfall on the operation of the ontology. The pitfall classification is an essential feature as not all pitfalls are equally relevant and important~\cite{PovedaVillaln2014}, as well as some modelling design choices may result in identified inconsistencies.

OntoClean is an independent method utilised to verify a correctly built ontology. Guarino and Welty introduced the OntoClean methodology in a series of conference papers in 2000~\cite{Guarino2000a, Guarino2000b, Guarino2000c}, which marked the beginning of a formal foundation for ontological analysis. OntoClean appropriates four notions to evaluate how \enquote*{clean} an ontology is: rigidity, identity, unity, and dependency. These four notions, adopted by knowledge-based systems as a means to model the world through \textit{meta-properties}, have existed in philosophy in an attempt to describe the universe in a formal, logical way since the early philosophical days of Aristotle~\cite{Guarino2009}. These form the basis of OntoClean to assist domain modellers in structuring a taxonomy in a logically correct ontology. See~\cite{Guarino2009, Guarino2000a, Guarino2000b} for further detail. The OntoClean method focuses on validating both the correctness and consistency of the taxonomic structure of an ontology~\cite{Guarino2009:ontologybook} and consists of two methodological steps
\begin{enumerate}
    \item Assignment of OntoClean meta-properties (rigidity, identity, unity and dependency) to the ontology classes.
    \item Verifying the ontology subsumption relationships.
\end{enumerate}
Evaluation with OntoClean in Prot\'eg\'e is achieved by~\cite{Mahlaza2019OntoCleanIO}
\begin{enumerate}
    \item Represent the ontology in the ABox from the TBox through punning the ontology.
    \item Assess the ontology and assign OntoClean meta-properties to each individual of the ABox.
    \item Discover inconsistencies in the taxonomic structure through a reasoner.
    \item Rectify inconsistencies by reassigning meta-properties to inconsistent classes.
\end{enumerate}
Note that ontology punning was conducted using the \textit{Cellfie} plugin to import and assign individuals with OntoClean meta-properties. Ontology evaluation with OntoClean can be time-intensive, primarily due to the time required to assign meta-properties to each class in ontology. Additionally, reasoning with logically expressive ontologies can present various issues to designers~\cite{ORE:2015}. We use the FaCT++ 1.6.5 reasoner with OntoClean, which is studied for ABox performance~\cite{Pan:2018} and evaluated against other state of the art reasoners such as Pellet and HermiT~\cite{ORE:2015}. The FaCT++ reasoner is tableaux-based designed for description logics and features optimisations to deliver performance enhancements over comparative reasoners~\cite{FaCT++:2006}.

\subsection{Designing an Ontology for Generalised Multi-Agent Teaming}\label{sec:design}

In this work, we address the lower-level of agent knowledge presented in Hepworth et al.~\cite{Hepworth2021:HST3}, including the agent logic services, through the design of an ontology. The HST3-Architecture identifies knowledge and model bases, swarm control and guidance functions, states knowledge base, planning functions, environment and context model, and reflection and learning functions as being contained at this level. Our ontology defines the function, context, and situations required to share semantic information at these service levels for a domain of interest as the information-sharing mechanism between humans and artificial agents. We posit that the services described in this architecture provide the fundamental elements required to share critical semantic information between agents. We assert that an ontology will help to deliver transparency through higher-level intent sharing and understanding for contexts and situations in HST missions (semantic information sharing), as described by Hepworth et al.~(2021)~\cite{Hepworth2021:HST3}. The key idea here is that a knowledge-grounded approach grounds the concepts with relations, providing context for an AI to understand and reason about the environment.

Knowledge elicitation co-occurred with the development of the ontology. The co-evolutionary approach enabled the top-level concepts from both teaming and shepherding to be identified quickly, contextualising the shepherding concepts within the teaming frame. We elicited knowledge through three different source categories, being
\begin{itemize}
    \item \textit{Academic and domain literature} containing the foundational concepts of shepherding and teaming systems. These covered shepherding from the modelling perspectives, see for instance~\cite{Abbass2020:ShepherdingUxVs, GIRARDI2008604, Baxter:HST-GO}, domain application perspectives through work such as Williams~(2007)~\cite{Williams2007:working}, and through to frameworks for HST, for example~\cite{Hepworth2021:HST3}.
    \item \textit{Domain expert exposure} through site visits to agricultural settings where biological shepherding occurs.
    \item \textit{Experimentation trials} to understand the interactions of biological and non-biological agents in the physical domain. Experimentation trials were approved under UNSW animal ethics committee ACEC No. 21/52B.
\end{itemize}

Our first knowledge elicitation task was to informally analyse academic and industry literature on shepherding to develop baseline understandings of the target domain for the shepherding application. This enabled the initial development of intermediate representations such as the glossary of terms, concepts and relation hierarchy trees. Initial exposure to domain experts was facilitated to understand further and conceptualise the domain, which aided to illuminate tacit knowledge previously undocumented. We then conducted a formal text analysis of the literature, identifying and confirming key concepts and lexicon, extracting necessary attributes, natural-language definitions, system rules and constraints, and agent properties.

Knowledge generated from the ontology design process for the function area of teaming and context and situation domain application areas (shepherding) was then synthesised into a consolidated concept hierarchy. Where synonyms conflicts occur\textemdash such as a \textit{flock} in shepherding or \textit{swarm} in HST literature, our design choices were guided by seeking to maximise the richness of the ontology for our objective to promote teaming between biological and artificial agents. In conjunction with this, redundant and duplicate concepts were removed, resulting in an irreducible set of concepts, relations, and individuals.

Shepherding is our context to facilitate enhanced teaming for situations where the control and guidance of multiple agents are paramount. Disparate use-cases that may not focus on this objective could require alternative teaming approaches, such as flocking, inspired by other biological systems. Determining the system objectives is essential to ensure the selected approach is appropriate for the desired outcome. Our knowledge elicitation process and subsequent information synthesis enable the generation of contextually-aware information for sharing between humans and artificial agents. The outlined knowledge contextualisation process provides the generalised method to \textit{build} additional application ontologies, be these for flocking or other teaming approaches, integrating new contexts and situations, creating a complete picture for a given function. The ontology helps us learn the system, translating from public understanding to formed knowledge.

\begin{figure*}[!h]
        \centering
        \includegraphics[width=0.95\textwidth]{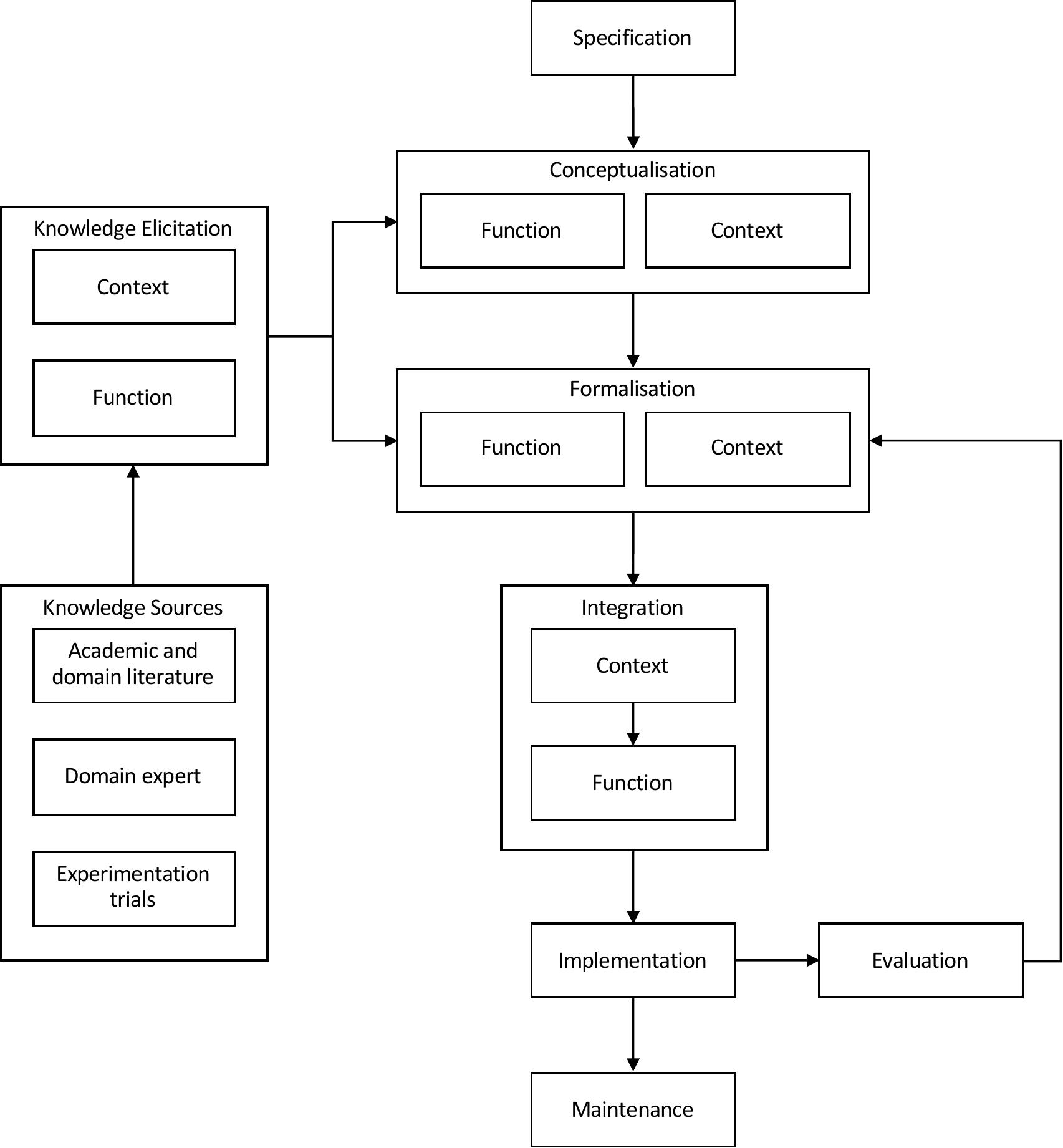}
        \caption{The design process used for Onto4MAT, based on IDEF5~\cite{IDEF5} and Methontology~\cite{Methontology:1997}. Our key difference is the fusion of a function (teaming) with context (shepherding) as the inspired representation approach. This diverges from classic formulations, typically more precise in their domain representation. However, they are seldom designed to support bi-directional understanding between agents and teams.}
        \label{fig:ontoDesignProcess}
    \end{figure*}

Figure~\ref{fig:ontoDesignProcess} describes our design process for Onto4MAT. Our choice of language remains consistent with the Methontology process; however, it diverges at the sub-module level for our application. We introduce the concepts of context and function as core elements of the design process. For our purpose, the function here is the desired end-state of teaming, where context is the method to achieve this. For the conceptualisation and formalisation phases, both the function and context are parallel processes that represent the processes outlined in IDEF5~\cite{IDEF5} and Methontology~\cite{Methontology:1997}. Our immediate departure occurs in integrating function and context, where a primary-secondary relationship is established. The purpose of this shift from parallel to serial information processing is to meet the purpose of the ontology. What we mean here is that the \textit{precise} representations developed during the conceptualisation phase are now transformed to achieve the intent. In Onto4MAT, the explicit representation of shepherding now forms the basis of the teaming function; however, it is not the primary representation desired. Shepherding enables teaming and is not the panacea for all teaming situations. This is important as context concepts are modified to nest with the function concepts prior to implementation. Our other departure here is how the evaluation is conducted, wherein lieu of being a continuous process, we focus our evaluation after the implementation phase. Post evaluation, an analysis of impacts for the context and function formalisation occur. This is important as we need to ensure that the function and context remain semantically consistent with the original formalisation, prior to integration revisions.

During the integration phase, the context is \textit{applied} to a function, with the function having precedence over the context. In an abstract setting, as additional contexts are included, different contexts can be applied to the function and only need to be integrated with the existing knowledge; however, still subordinate to the function. In the following section, we describe the upper ontology of Onto4MAT, which is the representation of the function for teaming, with sub ontologies being the relevant concepts of context from shepherding. As additional contexts are included, the upper ontology of Onto4MAT does not change; however, the richness of sub ontologies will.

\section{Ontology For Multi-Agent Teaming (Onto4MAT) Formal Specification}\label{sec:onto4mat}

In this section, we present the shepherding-based ontology, \textit{Onto4MAT}, as the output of our knowledge contextualisation method for teaming. We first provide an overview of the output ontology, describing the top-level concepts like classes, their relation to foundational teaming concepts and associated representations in the ontology and for this setting. We then discuss the upper ontology axioms and fundamental relations. We conclude this section by describing a use case for Onto4MAT, where the ontology is used in a scenario generation role to provide the intent, goals and constraints for a collection of agents to achieve a goal under the guidance of a shepherding-control agent. A human commands the swarm through interaction for a single intent. We first present in Table~\ref{table:ontoSummary} a summary of Onto4MAT key metrics.

    \begin{table}[]
    \centering
    \begin{tabular}{@{}ll@{}}
    \toprule
    \textbf{Metric}     & \textbf{Quantity} \\ \midrule
    Axiom               & 1,060             \\
    Logical Axiom Count & 562               \\
    Classes             & 167               \\
    Object Properties   & 57                \\
    Data Properties     & 16                \\
    Individuals         & 18                \\
    Primitive Classes   & 231               \\
    Defined Classes     & 30
    \end{tabular}
    \caption{Onto4MAT summary statistics.}
    \label{table:ontoSummary}
    \end{table}

\subsection{Onto4MAT Upper Ontology Classes}

\begin{figure}[!h]
     \centering
     \begin{subfigure}[b]{0.3\textwidth}
         \centering
         \includegraphics[width=\textwidth]{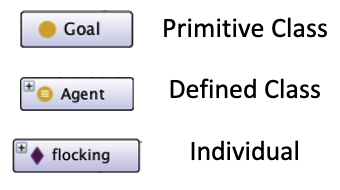}
         \caption{Classes and Individuals.}
         \label{fig:explained1}
     \end{subfigure}
     \hfill
     \begin{subfigure}[b]{0.4\textwidth}
         \centering
         \includegraphics[width=\textwidth]{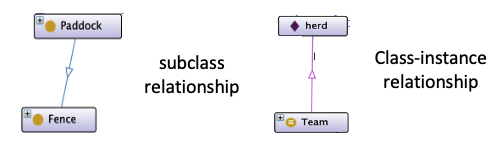}
         \caption{Class-Instance and Subclass relationships.}
         \label{fig:explained3}
     \end{subfigure}
     \hfill
     \begin{subfigure}[b]{0.5\textwidth}
         \centering
         \includegraphics[width=\textwidth]{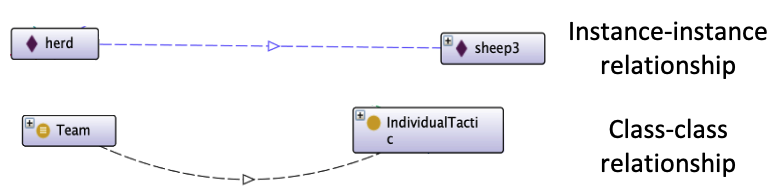}
         \caption{Class-Class and Instance-Instance relationships.}
         \label{fig:explained2}
     \end{subfigure}
        \caption{Explanation of figures contained in Section~\ref{sec:onto4mat}.}
        \label{fig:explainedFigures}
    \end{figure}

\begin{figure*}[!h]
        \centering
        \includegraphics[width=0.95\textwidth]{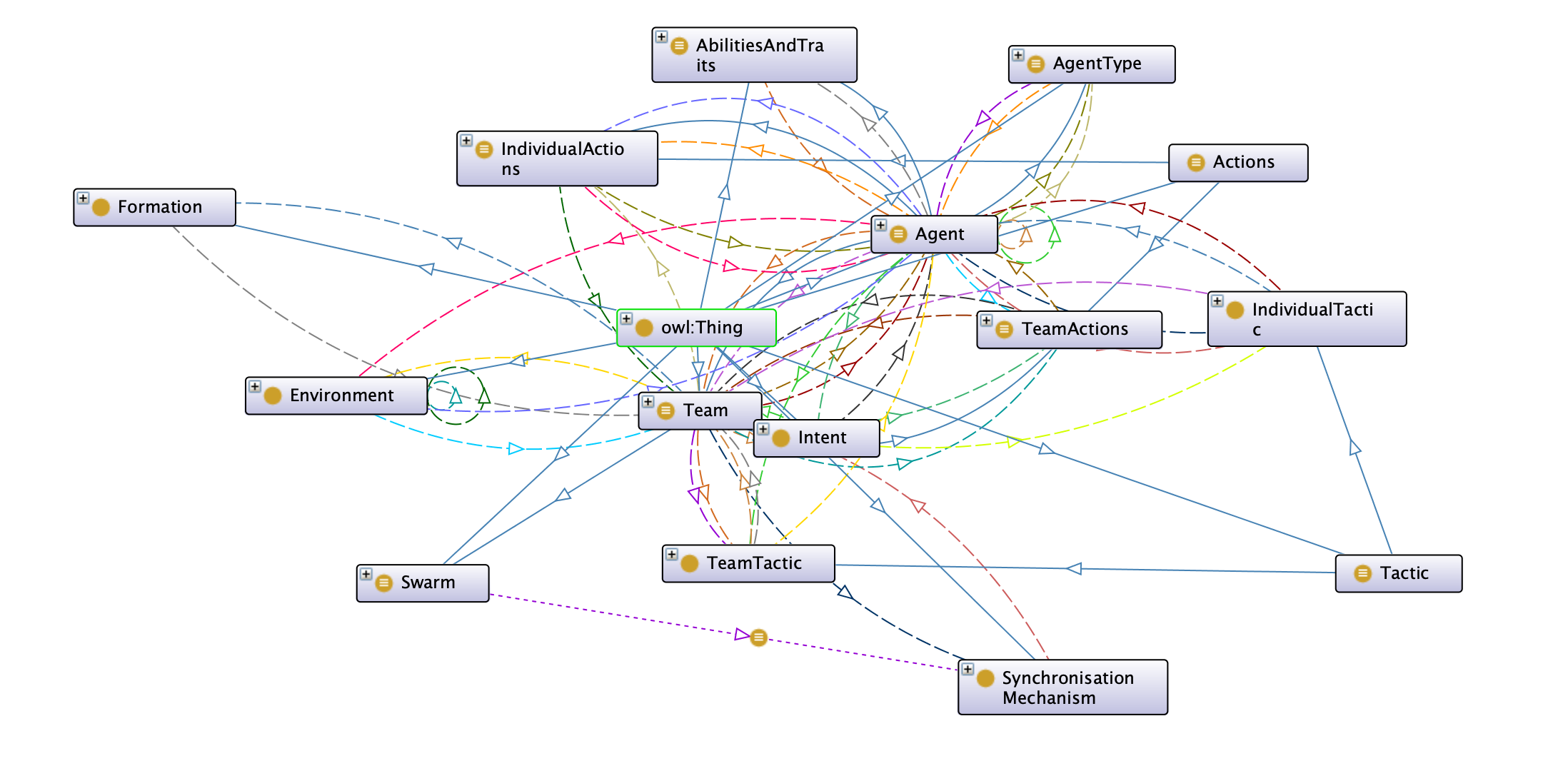}
        \caption{Onto4MAT upper ontology depicting asserted classes and their relations for generalised multi-agent teaming.}
        \label{fig:upperOntology}
    \end{figure*}

    The upper ontology of Onto4MAT contains the primary concepts required to share high-level semantic understanding among agents in a team. These general concepts represent the result of the contextualisation method and focus on the agents, their type, abilities and traits, the actions and tactics they can perform, and intent. Contained at the same hierarchy level are the central concepts of the team, which, when combined with a synchronisation mechanism, intent, collective actions, tactics, and a formation, form the basis of concepts for a swarm. The majority of concepts at this level are given as defined classes by the \texttt{EquivalentTo} expression, which indicates that \enquote{two concepts have the same intensional meaning}~\cite{w3:owl}. Figure~\ref{fig:upperOntology} depicts the Onto4MAT upper ontology, with semantic relationships (object properties) illuminating the \texttt{domain} and \texttt{range}, consistent with~\cite{w3:owl}. With swarm shepherding being the primary use-case application of this ontology, the upper-level ontology classes are concept-consistent with the concepts contained in Williams~(2007)~\cite{Williams2007:working}, and Abbass and Hunjet~(2021)~\cite{Abbass2020:ShepherdingUxVs}. Before discussing Onto4MAT, we provide the following description of the various object types and relationships, defined through symbols in Figure~\ref{fig:explainedFigures}.

    \subsubsection{Agent}

\begin{figure*}[!h]
        \centering
        \includegraphics[width=0.95\textwidth]{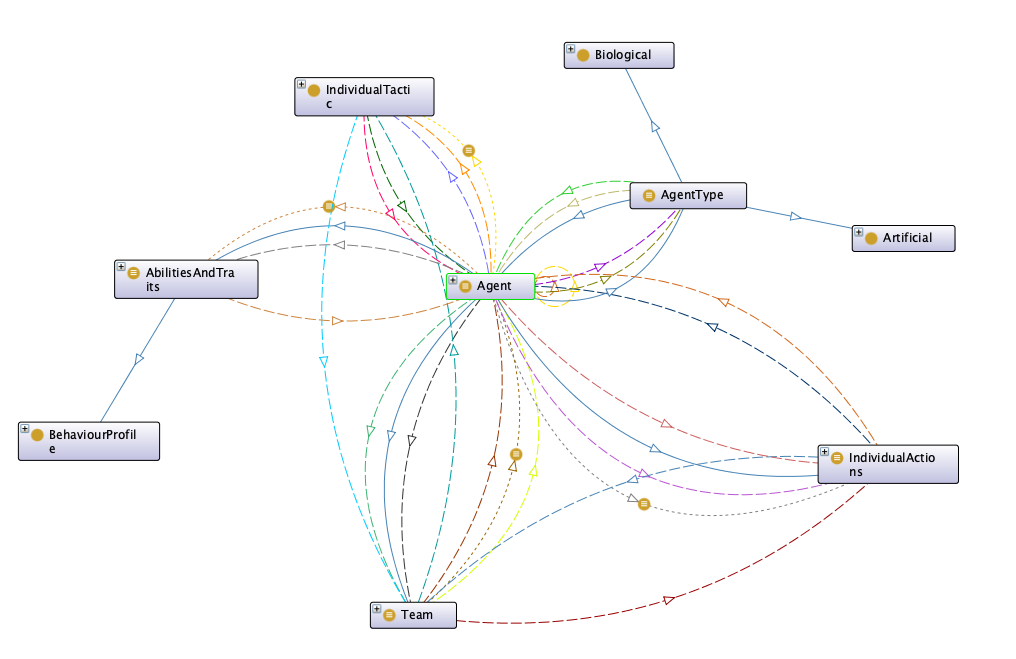}
        \caption{The class \texttt{Agent}, its sub-class hierarchy and related upper-ontology classes, defining the types of agent's within a shepherding system.}
        \label{fig:agent}
    \end{figure*}

    The \texttt{Agent} class focuses on the representation of an agent fulfilling one of three primary types in a shepherding situation, being that of the shepherd (commander, $\zeta$), the sheepdog (shepherding control agent, $\beta$), and the sheep (team member, $\pi$)~\cite{Hepworth2021:ARS}. Note that we include the ability to define an artificial type of agent here, available to fulfil any of the three primary types. We define an \texttt{Agent} based on the \texttt{AgentType}, which relates the type to the \texttt{AbilitiesAndTraits}, \texttt{IndividualActions}, and \texttt{IndividualTactics} to the set of available instances in those classes. The \texttt{Agent} class is related to the defined class \texttt{Team}. As with the majority of classes, there are multiple relationships defined between \texttt{Agent} and \texttt{Team} to account for not only the three primary types above but also the resultant interaction space. The hierarchical structure of the \texttt{Agent} class are given in Figure~\ref{fig:agent}, which also depicts the sub-class relations.

    \subsubsection{Abilities and Traits}

\begin{figure*}[!h]
        \centering
        \includegraphics[width=0.95\textwidth]{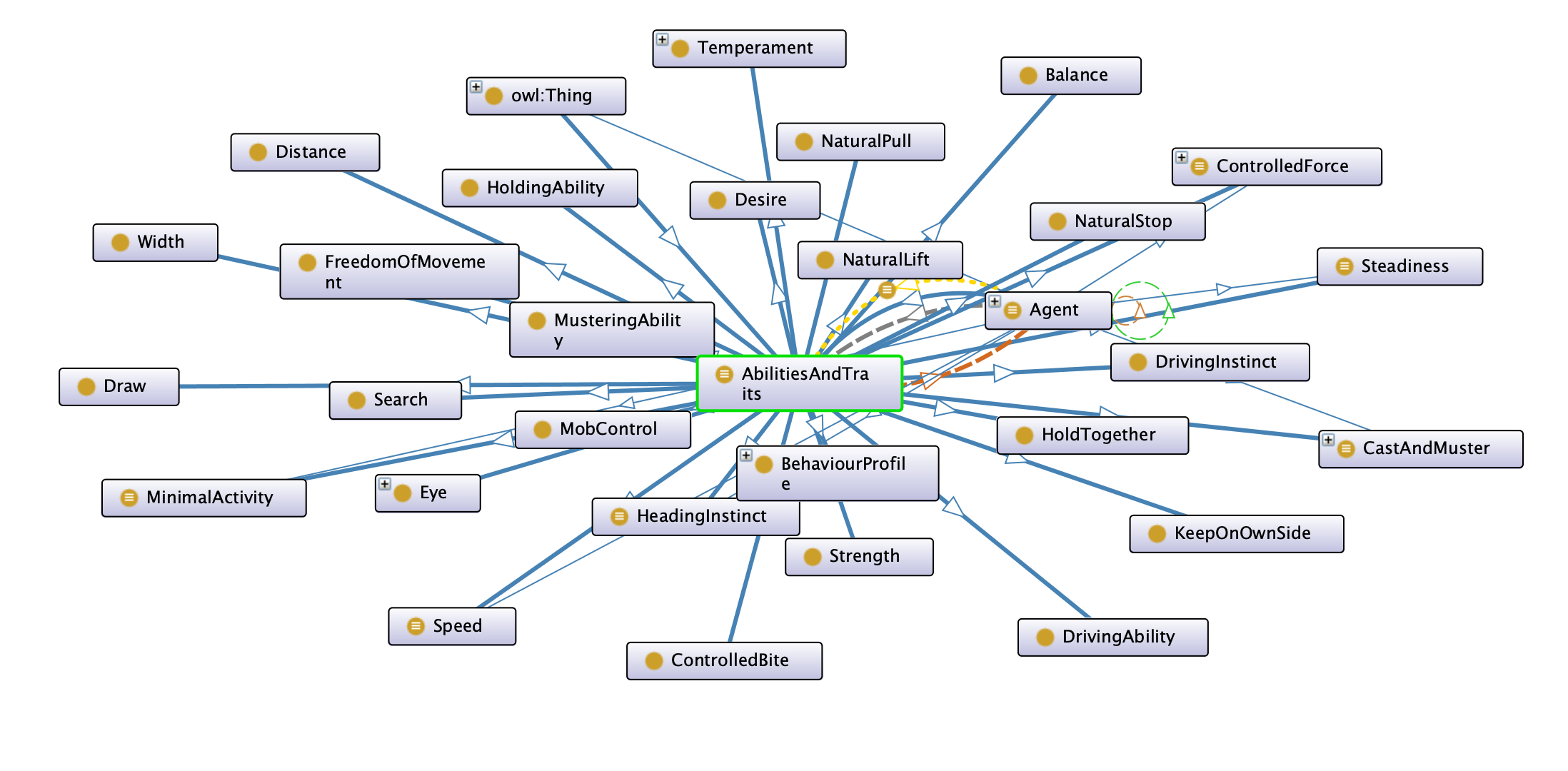}
        \caption{The \texttt{AbilitiesAndTraits} class depicting the range of agent abilities and traits for shepherding-based teaming. Note the existence of both primitive (yellow circle) and defined (yellow circle with white lines) classes for abilities and traits, indicating the interdependent class descriptions.}
        \label{fig:abilitiesTraits}
    \end{figure*}

    In the Onto4MAT, agents may posses unique, non-homogeneous abilities and traits, being attributes of the agent. The abilities and traits outlined in Figure~\ref{fig:abilitiesTraits} depict a shallow hierarchy structure, with the majority of concepts sub-classes of \texttt{AbilitiesAndTraits}. As a defined class, \texttt{AbilitiesAndTraits} is \texttt{EquivalentTo} an \texttt{Agent} who hold some subset of the possible abilities and traits. The broad range of abilities and traits defined facilitates the heterogeneous parameterisation of agents, reflecting real-world observations~\cite{Hepworth2020:Footprints}. This enables the inference of behaviour profiles to classify agents and discover their role within social and organisational hierarchies, which is an active area of research across multiple disciplines~\cite{Basak2021:leaderFollower, garland:2018, jiang2017:flocking}. When attributed to an agent, each ability and trait provides a unique capability to an agent.

    \subsubsection{Team}

\begin{figure*}[!h]
        \centering
        \includegraphics[width=0.95\textwidth]{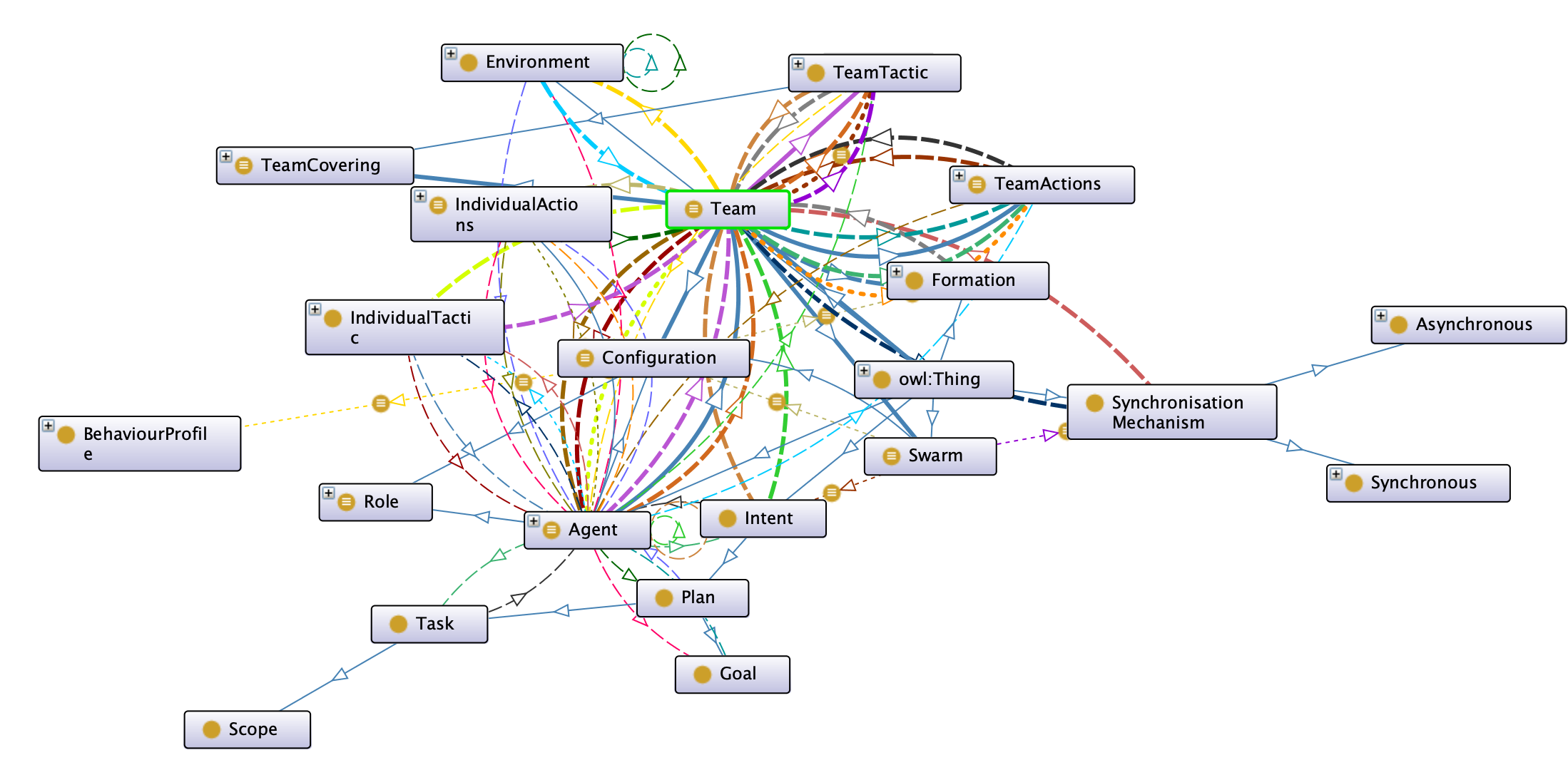}
        \caption{The \texttt{Team} class represents a rich concept which is essential to both the shepherd (commander) and sheepdog (shepherding control agent) to achieve their desired outcomes.}
        \label{fig:team}
    \end{figure*}

    A team within Onto4MAT represents a collection of agents who undertake tactics or actions. As a defined class, \texttt{Team} is given as \texttt{EquivalentTo} an \texttt{Agent}, at least two agent's conduct either \texttt{TeamActions} or \texttt{TeamTactic}. The team is a foundational concept that links agents to a swarm through an organisation. Abbass and Hunjet~(2021~\cite{Abbass2020:ShepherdingUxVs}) define a team as \enquote{a group of organised individuals joined together to execute team-level tactics and actions}. The \texttt{Team} class, its relations and sub-classes are given in Figure~\ref{fig:team}, which depict the key relationships to intent, configuration (being collective formation and agent role), and swarms. The \texttt{Agent} and \texttt{Team} classes represent the vast majority of relation uses within the ontology, representing nearly half of all total uses of upper-ontology concepts.

    \subsubsection{Intent}

    Intent, $\mathfrak{I}$, and its representation are fundamental to the selection of shepherding as our approach to design a generalised teaming ontology. An intent is not only a situation constraint but too delivers a unique context through information, in this instance from the shepherd through the sheepdog to the sheep \textemdash from commander to the shepherding control agent and eventually to team members. When an intent is sufficiently and successfully transferred from one agent to another, intent enables the receiving agent to develop their sequence of actions to achieve an outcome. From a recognition perspective, observing actions and tactics may enable the inference of an intent to generate higher situational and context awareness~\cite{Abbass2020:ShepherdingUxVs, Hepworth2021:ARS}. In Onto4MAT, the class \texttt{Intent} is a primitive which provides rich semantics to the upper-level ontology. In this setting, intent consists of a plan, goal task and scope as the defined set of elements for the shepherd agent to provide the necessary context to the shepherding control agent to actuate in the environment.

    \begin{figure*}[!h]
        \centering
        \includegraphics[width=0.65\textwidth, height=0.5\textheight]{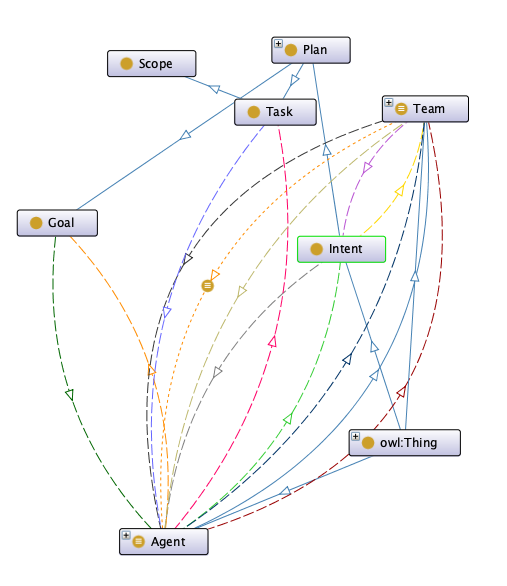}
        \caption{The \texttt{Intent} class relates the concepts of \texttt{Agent}, \texttt{Team}, and \texttt{Swarm} to the Shepherd and Sheepdog agent types, enabling the Sheepdog to conduct missions.}
        \label{fig:intent}
    \end{figure*}

    \subsubsection{Actions}

    Actions, also known as behaviours ($\Sigma =\{\sigma_1, \sigma_2, \dots, \sigma_K\}$), in Onto4MAT are inspired by the definition given by Liu et al.~(2016)~\cite{Liu:2016:AA:2894223.2894265}, given as \enquote{primitives that fulfil a function or simple purpose}. We define actions at one of two levels as individual or collective. Individual actions consist of behaviours that an agent may conduct, whereas collective behaviours are the behaviours that a team may conduct. Actions fulfil two key roles for agents in the ontology. The first is the set of behaviours that may be conducted, with each agent type possessing a subset of actions they may select to do. The second is for an agent, or external observer, to recognise the current situation of an agent. Once classified and labelled, these semantic labels allow us to understand the state of an agent. When combined with observed or deduced contexts, an observer can ascertain the state of an agent or team. For the shepherding agent, understanding the team's situation they are tasked to control is a context that conditions their action selection to achieve the objective. As with the \texttt{AbilitiesAndTraits} class, the \texttt{Actions} class is possesses a predominately shallow hierarchy structure. \texttt{AbilitiesAndTraits} is defined as an \texttt{Agent} who conducts \texttt{IndividualActions} or \texttt{Team} who conducts \texttt{TeamActions}.

    \begin{figure*}[!h]
        \centering
        \includegraphics[width=0.95\textwidth]{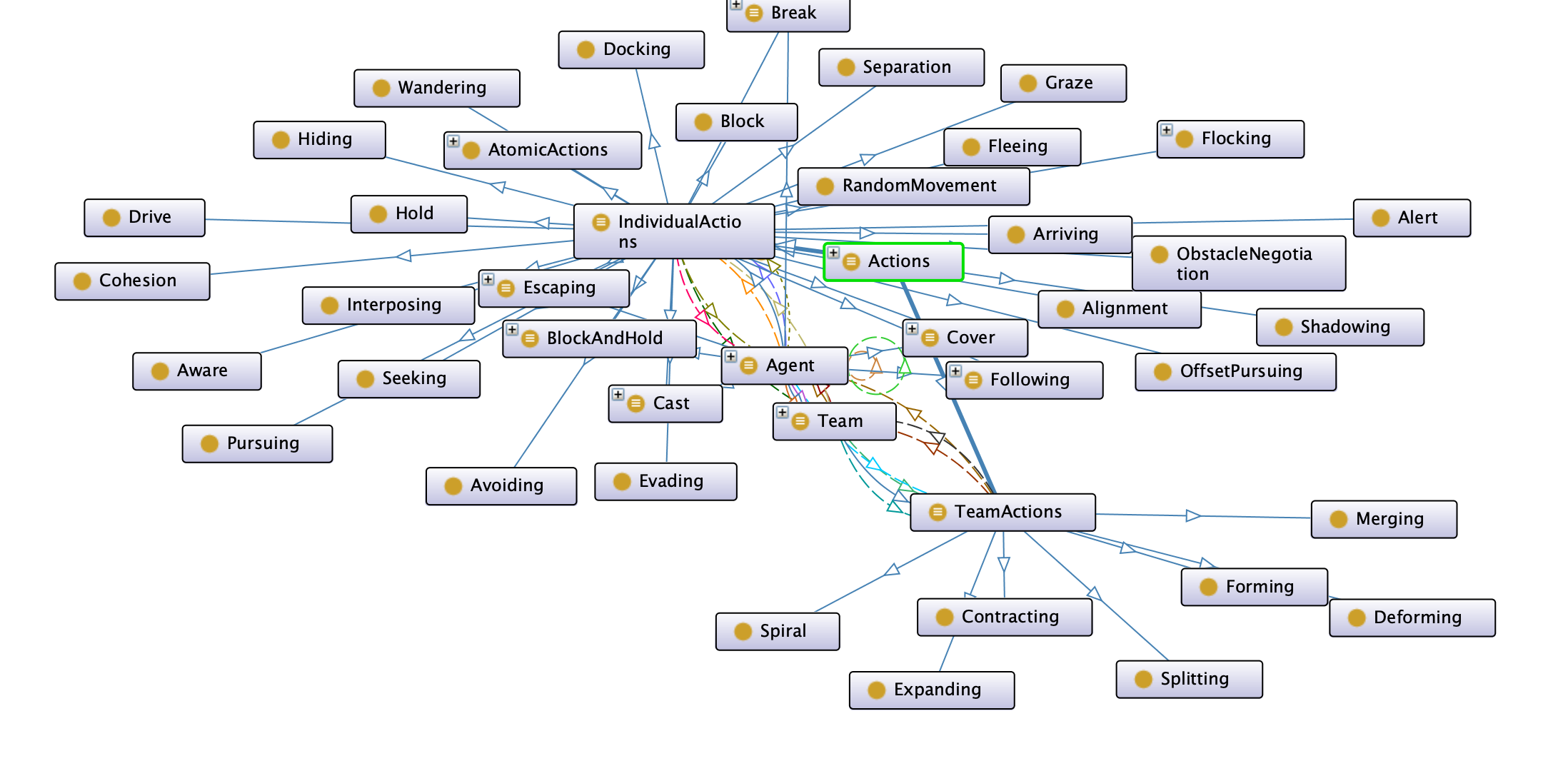}
        \caption{The \texttt{Actions} class defines individual and collective behaviours which may be conducted or recognised in the shepherding system.}
        \label{fig:action}
    \end{figure*}

    \subsubsection{Tactic}

    The boundary we use to define between actions and tactics, $\Psi =\{\psi_1, \psi_2, \dots, \psi_K\}$, is inspired by the work of Ikizler and Forsyth~(2007)~\cite{4270193}, who state \enquote{we distinguish between short-timescale representation (acts); medium timescale actions, like walking, running, jumping, standing, waving, whose temporal extent can be short (but may be long) and are typically composites of multiple acts; and long-timescale activities, which are complex composites of actions.} In shepherding, temporal separations may not always be appropriate as tactics may be short and actions may be prolonged. We focus on the complexity composites portion of Ikizler and Forsyth's definition, with this as our guiding principle for differentiating actions and tactics. Given this, in our setting, represent \enquote{an organised set of team actions to achieve an intent or higher-order effect}~\cite{Abbass2020:ShepherdingUxVs}. Abbass and Hunjet (2021) include the notion of an objective or effect as the output of the tactic. The objective to be achieved or effect to be generated in Onto4MAT is intent.

    \begin{figure*}[!h]
        \centering
        \includegraphics[width=0.95\textwidth]{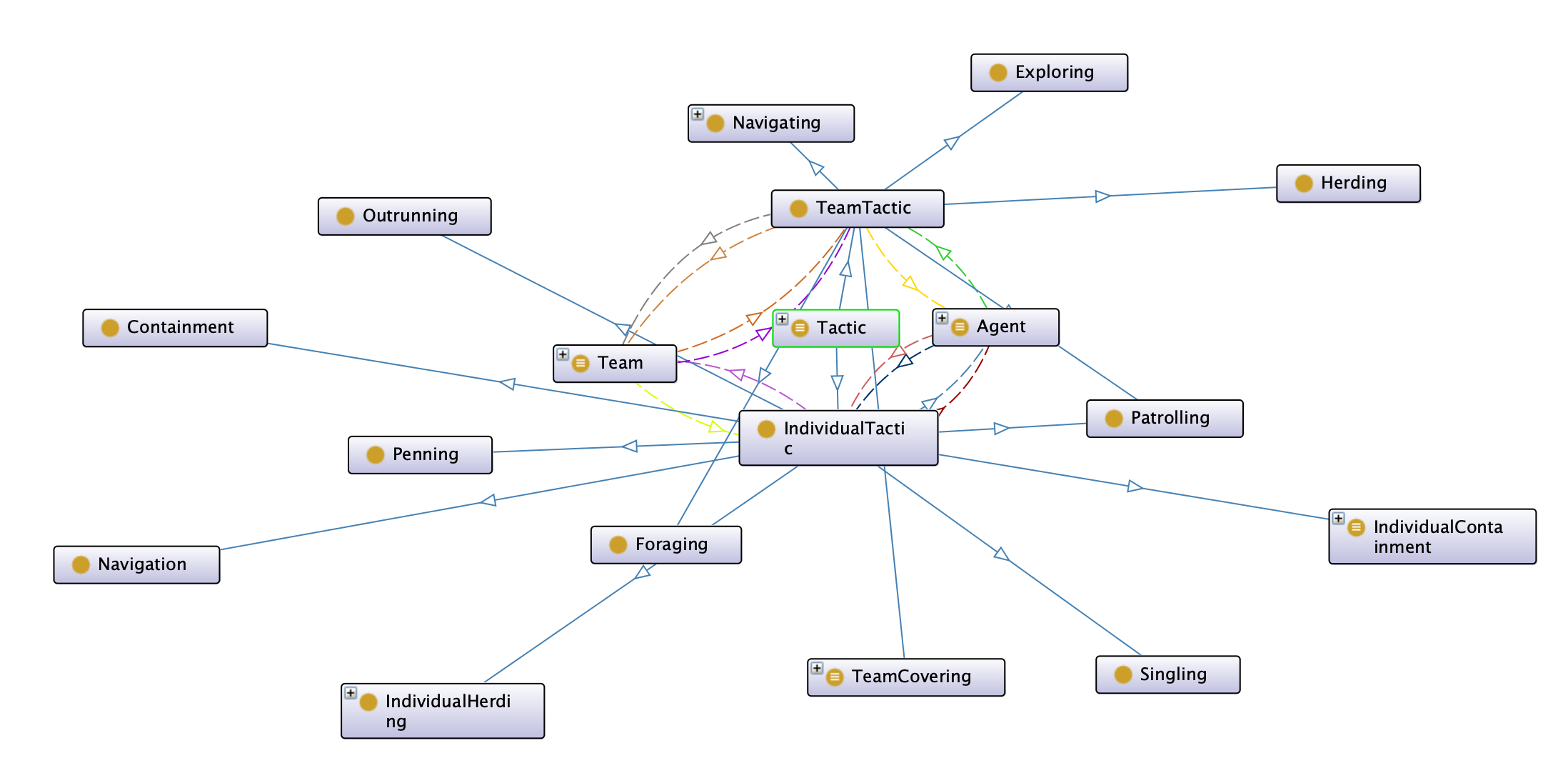}
        \caption{The \texttt{Tactic} class contains both individual and collective tactics~\textemdash composite sequences of actions~\textemdash which are conducted and observed in shepherding systems.}
        \label{fig:tactic}
    \end{figure*}

    \subsubsection{Swarm}

\begin{figure*}[!h]
        \centering
        \includegraphics[width=0.95\textwidth]{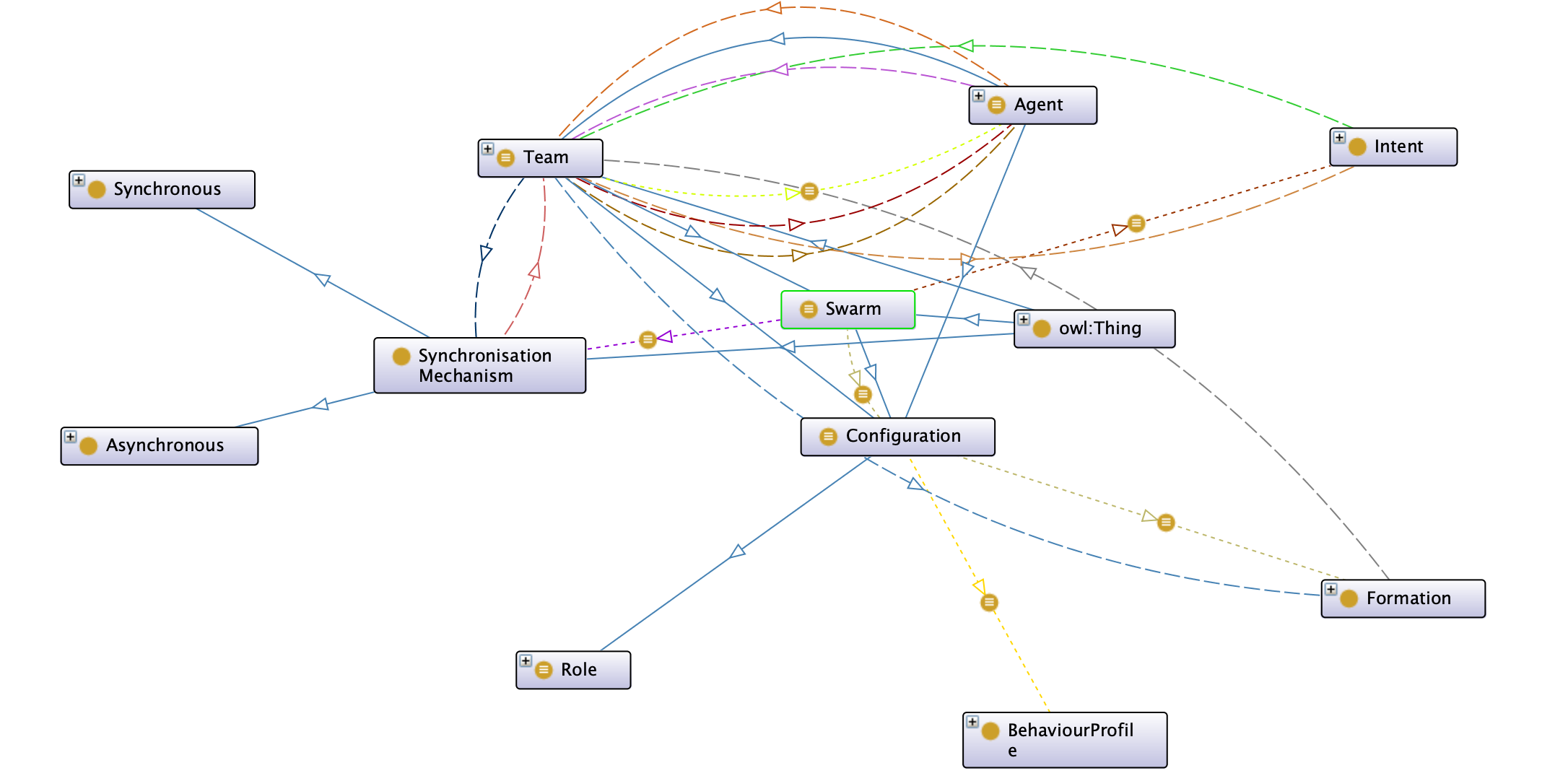}
        \caption{The \texttt{Swarm} class relates upper-ontology concepts together, enabling the achievement of intents by swarms guided by a shepherding control agent.}
        \label{fig:swarm}
    \end{figure*}

    A swarm, $\Pi = \{\pi_1, \pi_2, \dots, \pi_N\}$ is defined by Abbass and Hunjet~(2021) as a \enquote{team with actions of the individuals that are aligned spatially and/or temporally using a synchronisation strategy}~\cite{Abbass2020:ShepherdingUxVs}. We extend the two primary concepts here of a team and synchronisation strategy with the inclusion of intent at the swarm level and the notion of configuration. Configuration in this setting is more than the spatial swarm formation, including agent type and role concepts. Agent type is important here to define an agent's capabilities (abilities and traits), with role important to define an agent's social and organisational position in a swarm. This provides an increased understanding of the swarm and offers novel ways to identify influence, leadership, and other traits in the swarm, both for recognition and control. The \texttt{Swarm} class relates \texttt{Team} and \texttt{Intent} with \texttt{Configuration}, being a function of agent type and role, and team formation, and \texttt{SynchronisationMechanism}, which is depicted in Figure~\ref{fig:swarm}.

    \subsubsection{Environment}

\begin{figure*}[!h]
        \centering
        \includegraphics[width=0.95\textwidth]{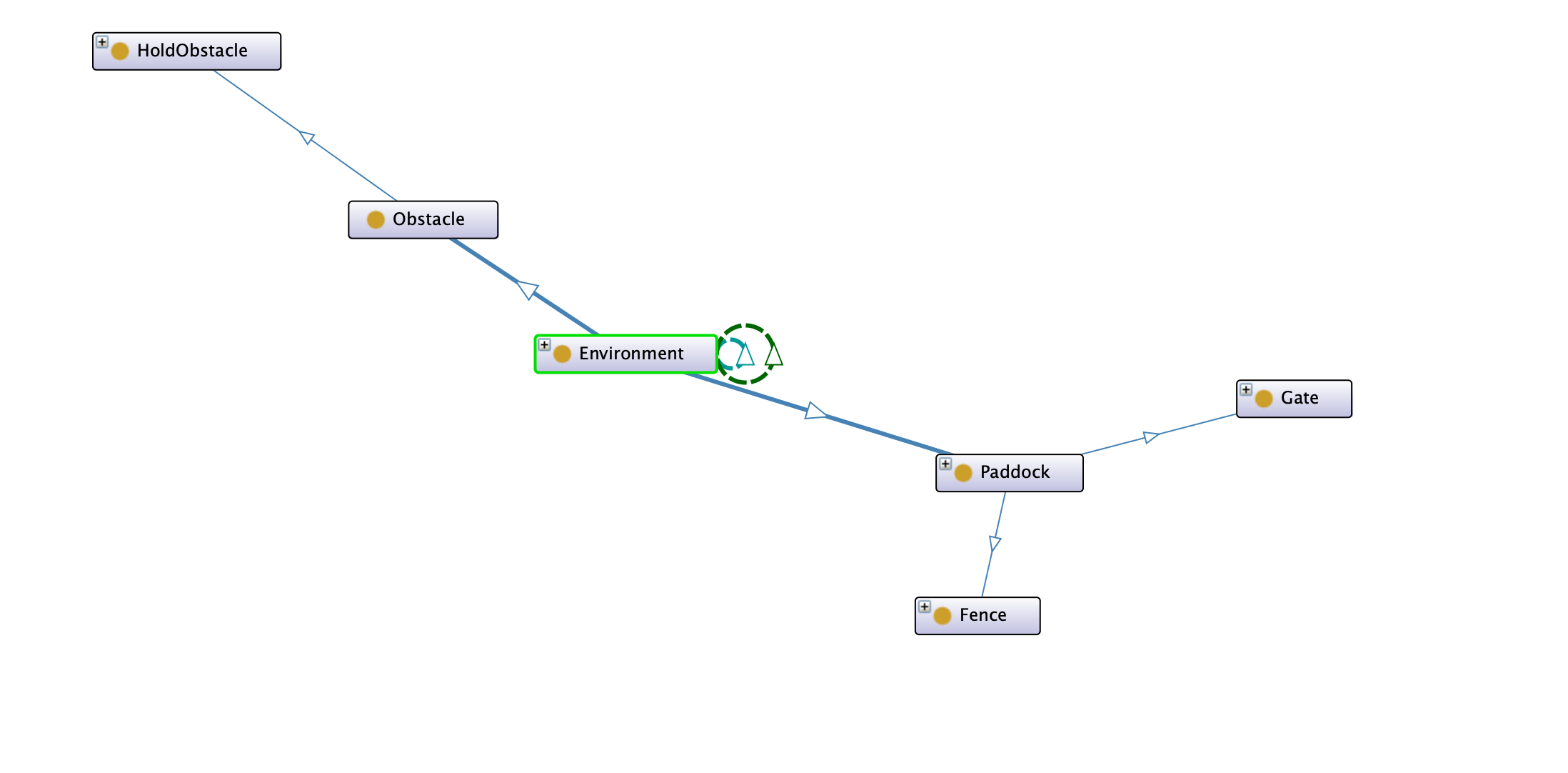}
        \caption{The \texttt{Environment} lexicon represents the shepherding environment; however, the concepts contained generalise beyond this setting.}
        \label{fig:environment}
    \end{figure*}

    The \texttt{Environment} class, $\mathcal{E}$, contains the information set required for the previously discussed upper-level ontology concepts to be realised. While faithfully representing the lexicon of the shepherding domain, the selected semantics generalise beyond an initial inspection, with the environment being the location where all agents are situated (context). The \texttt{Paddock} sub-class represents the conceptual location of agents, be it physical, information, or another representation. The \texttt{Obstacle} sub-class reflects a constraint on the achievement of an intent appropriate for both the physical and information domains. Figure~\ref{fig:environment} reflects this scope, with physical shepherding domain sub-classes present to contextualise this discussion.

\subsection{Ontology Axioms Overview}

Staab and Maedche (2000) outline seven higher-level categories of axiom types, defining axioms as \enquote{complex objects that refer to concepts and relations}~(p.6)~\cite{Staab:2000}. Attributes of Onto4MAT classes and instances are described through axioms, primarily the object and data properties. Object properties relate classes to individuals, where data properties define the location and what type of data may be asserted to the ontology. We define axioms that specify relationships through predicate logic, associating usage requirements and characteristics. The use of axioms as logical information about classes and properties provides a way to reason about the world and imposes constraints on classes and their individuals. Semantics are derived by inferring additional information based on the assertion of detailed data.

\begin{figure*}
    \centering
    \includegraphics[width=0.75\textwidth]{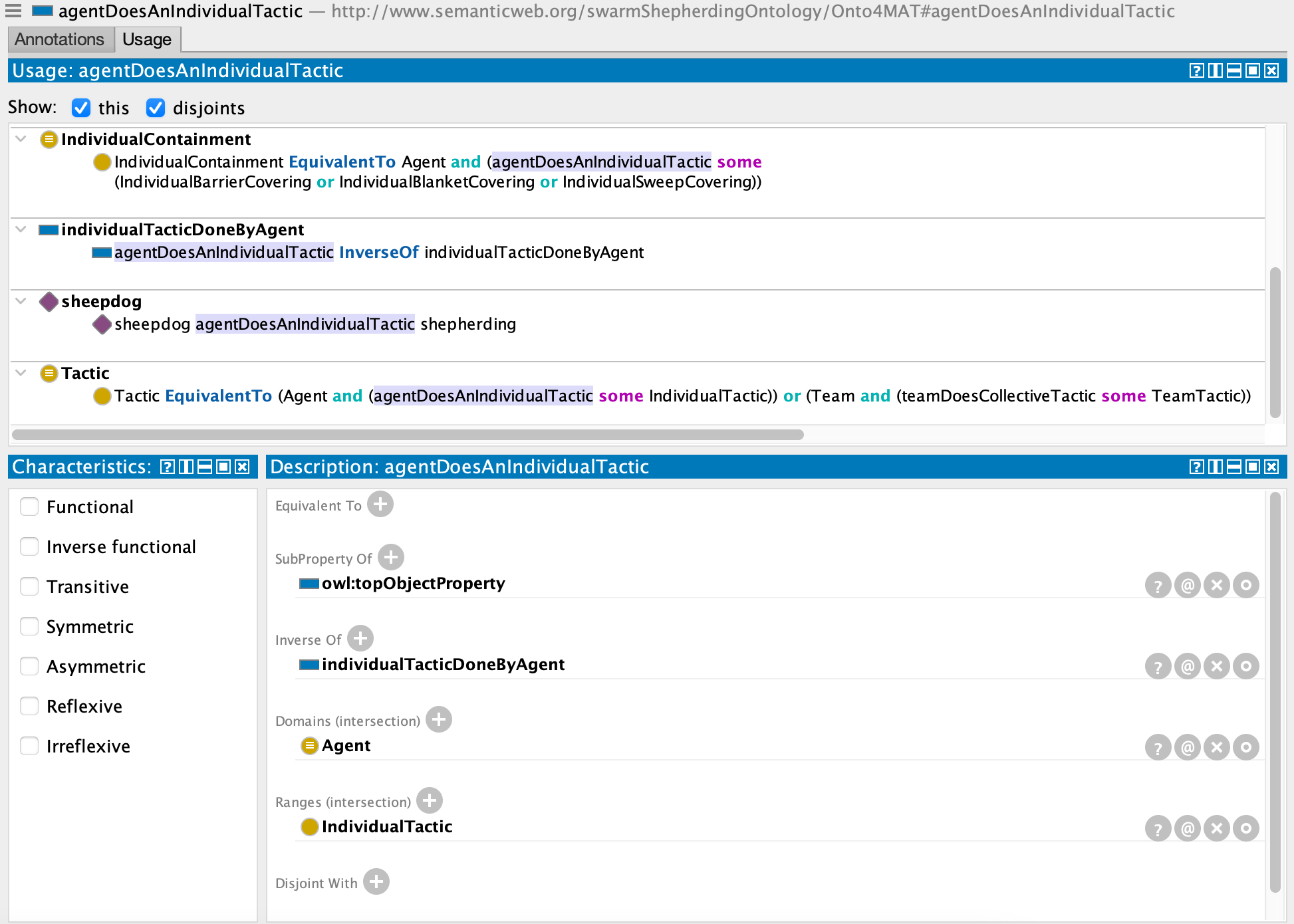}
    \caption{Example of the object property \texttt{agentDoesAnIndividualTactic} and a selection of uses within Onto4MAT.}
    \label{fig:objectProperties}
\end{figure*}

The object properties of Onto4MAT provide logical relations to enable semantic interpretation, of which 57 are defined to enable this. Object properties are broken into seven functional groups being, \texttt{controls}, \texttt{does}, \texttt{partOf}, \texttt{has}, \texttt{is}, \texttt{affects}, and \texttt{influences}. Each abstract category contains the set of object properties for that type, following a naming convention which conveys semantic meaning between the classes and individuals related through the property. For example, the object property \texttt{teamHasAgent} has the domain \texttt{Team} and range \texttt{Agent}. 21 uses are recorded in the ontology, including for the definition of a team, which is given as \texttt{Agent and ((teamDoesCollectiveAction some TeamActions) and\newline (teamDoesCollectiveTactic some TeamTactic) and (teamHasAgent min 2 Agent))}. The object property \texttt{agentDoesAnIndividualTactic} has domain \texttt{Agent} and range \texttt{IndividualTactic}, recording 21 uses within the ontology. This object property is additional used to relate individuals, for example \texttt{sheepdog agentDoesAnIndividualTactic shepherding}, which is depicted along with other uses in Figure~\ref{fig:objectProperties}

\begin{figure*}[!h]
        \centering
        \includegraphics[width=0.95\textwidth]{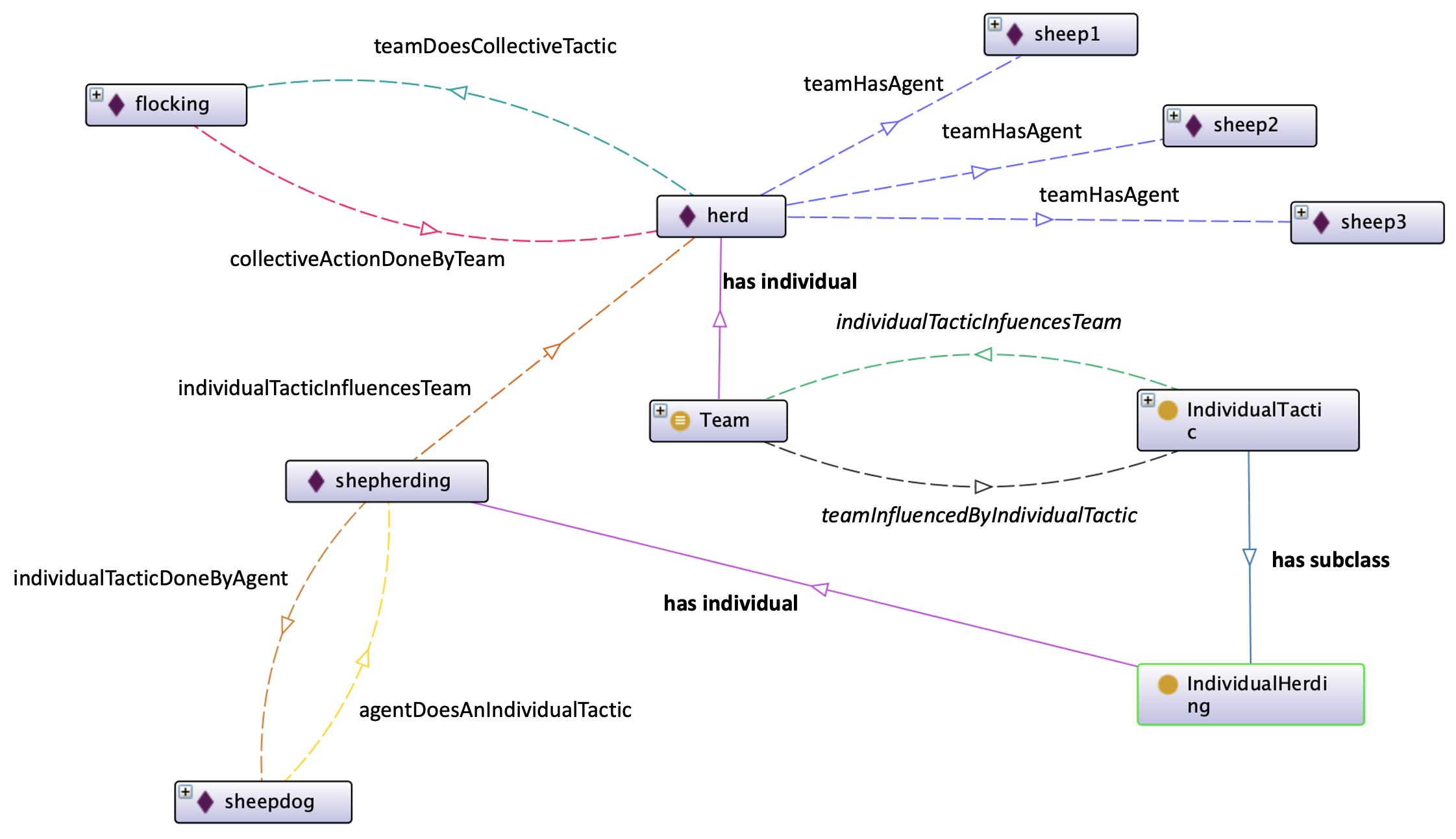}
        \caption{Axiom structure for individuals, class \texttt{domain} and \texttt{range}, and datatype properties for an example scenario of the shepherding control agent herding a swarm.}
        \label{fig:axioms}
    \end{figure*}

Figure~\ref{fig:axioms} depicts a subset of Onto4MAT classes and individuals, related through descriptive axioms. In this selected subset, we see the individual \texttt{sheepdog} is related to the individual \texttt{shepherding} through the object property \texttt{agentDoesAnIndividualTactic}. We can infer from this relationship that the sheepdog is an agent who does the shepherding tactic. \texttt{shepherding} is related to the individual \texttt{herd} by the object property \texttt{individualTacticInfluencesTeam}, indicating that the shepherding tactic has influence on the \texttt{Team} class individual,  \texttt{herd}. From the perspective of \texttt{herd}, we observe that it has team members of \texttt{sheep1, sheep2, sheep3}, who undertake the collective action of \texttt{flocking}. While a small example, this demonstrates the power of the ontology to enable the sheepdog agent to select actions and tactics which influence a team to achieve the intent of a commander \textemdash a shepherding-based approach to teaming.

\section{Ontology-Guided Experimentation}

Designed to enable teaming between agents, we demonstrate the use of Onto4MAT through an application of the ontology to enable intent delivery and action production in a simulated shepherding environment. During the initialisation of a shepherding simulator, many parameters are needed for the agent and the environment, such as an agent's behaviours, locations, sensors, goals, abilities and traits, interaction ranges and environmental dimensions. Typically, these are hard-coded into the simulation and the behaviours available to an agent. Onto4MAT offers several opportunities to provide a generic agent architecture to model the problem and concept context definitions for shepherding. The ontology is read and understood to do shepherding tasks. Another role of Onto4MAT in simulation settings is to translate human intent from natural language to an interpretable state-space format for an agent to conduct shepherding. This translation provides the concepts and logic necessary to parameterise a simulator in place of hard-coded elements. A further opportunity is for a human to query an agent through the ontology, delivering a meaningful interaction.

The focus of our simulation in this paper is scenario generation. The use of ontologies to support scenario generation is established in the literature across many fields. Jeong et al.~(2016)~\cite{Jeong:2016} propose a system to support military decision making, to \enquote{minimise the damage of our forces and make an optimal decision.} Sorokine et al.~(2011)~\cite{osti_1052239} propose an ontology-driven information system which scales the generation of \enquote{a much larger number of specific scenarios based on limited user-supplied information} for interactive simulations. Filho et al.~(2015)~\cite{FilhoAV15} design an ontology-guided scenario generation operator training tool, noting that \enquote{scenario development based on ontology provides a common language among stakeholders favouring information sharing and re-use.}

The primary perspective within the literature is of a user interacting with an ontology-based system to support human-based decisions and understanding. The user could be considered distinct from the system in these settings, interacting with it uni-directional. The view in our work is that the user is part of the system, delivering an intent for bi-directional agent interactions. Through this lens, a human shepherd provides an intent to a shepherding control agent through the Onto4MAT. The ontology contextualises the intent of the human for the shepherding control agent, reasoning to infer a goal location ($P_G$), the available set of tactics ($\psi$), the available set of behaviours ($\sigma$), and the agent(s) to focus on in the swarm. Once delivered and understood by the shepherding control agent, the task is conducted to achieve the commander's intent. Formally, we state $\zeta_{\mathfrak{I}}\rightarrow\beta = f(P_G, \Psi, \Sigma, \Pi)$.

A human assumes the role of the shepherd and interacts with the sheepdog through the ontology, where a dynamic query is generated on the ontology to develop the necessary detail to execute the simulation. The function of this interaction is for the shepherd to deliver an intent. The query is generated, which finds the closest $\psi_\beta$ to the given $\mathfrak{I}$; for this, we assume a $1:1$ mapping between $\mathfrak{I}$ and $\psi$. The ontology reasoner infers possible $\sigma_\beta$ to achieve $\psi$. In this example, the shepherd inputs $\mathfrak{I}=$\texttt{mustering}, which returns $\psi_\beta=$\texttt{mustering} as the tactic, with allowed behaviours of $\sigma_\beta=\{$\texttt{collect}, \texttt{drive}$\}$, inferred through the object property \texttt{taskForAgent}, connecting $\psi$ and $\sigma$.

Success criteria and mission constraints, such as a goal location and the size of $\Pi$ are also defined. Prior to the execution of the simulation, an \textit{a priori} explanation is generated. This synthesises the shepherd's input and the contexts and situations inferred by the ontology. The human shepherd must concur with the briefed mission prior to its conduct. Once the shepherd's approval is given, the ontology instantiates the simulation with the required parameters and commences the task. This simulated example describes the bi-directional flow of semantic information, demonstrating the capability of the ontology to share semantic-level information between agents to the team.

\begin{figure*}[!h]
    \centering
    \includegraphics[width=0.75\textwidth]{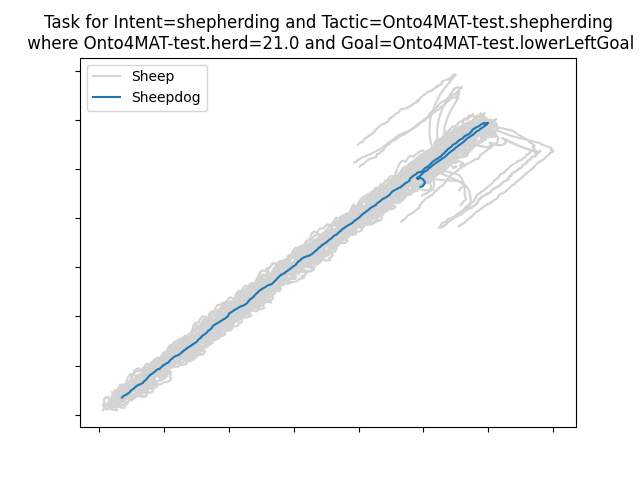}
    \caption{Example simulation output depicting the path of all agents and inferences of Onto4MAT to generate the parameterised scenario.}
    \label{fig:simOut}
\end{figure*}

\section{Evaluation}\label{sec:eval}

Our method consists of eight stages, with two iterations of four distinct elements, comprising three verification methods and one validation method. The first evaluation stage combines verification methods 1 and 2, OntoClean and OOPS!, which diagnose and clean the ontology technically prior to verification method 3\textemdash semantic and domain relationship verification by a domain expert in stage 2. In stage 3, we re-diagnose and clean the ontology to ensure any modifications made during stage 2 are technically sound. At this point, we assume the ontology has been verified and is ready for validation. Stage 4, the final stage of the first iteration, is validation method 1\textemdash end-user validation. Upon completion of the first iteration, the feedback obtained through end-user validation has the potential to influence properties within the ontology, such as classes, instances, objects, relations, and properties. Consequently, it is essential to re-verify and validate the ontology to ensure no errors have been introduced\textemdash iteration two of the four stages $i=1$. For this paper, we limit the ontology evaluation to stage 1 iteration 1 (methods 1 and 2) to maintain focus on delivering the formal knowledge representation of a generalised HST ontology.

\begin{figure*}
    \centering
    \includegraphics[width=\textwidth]{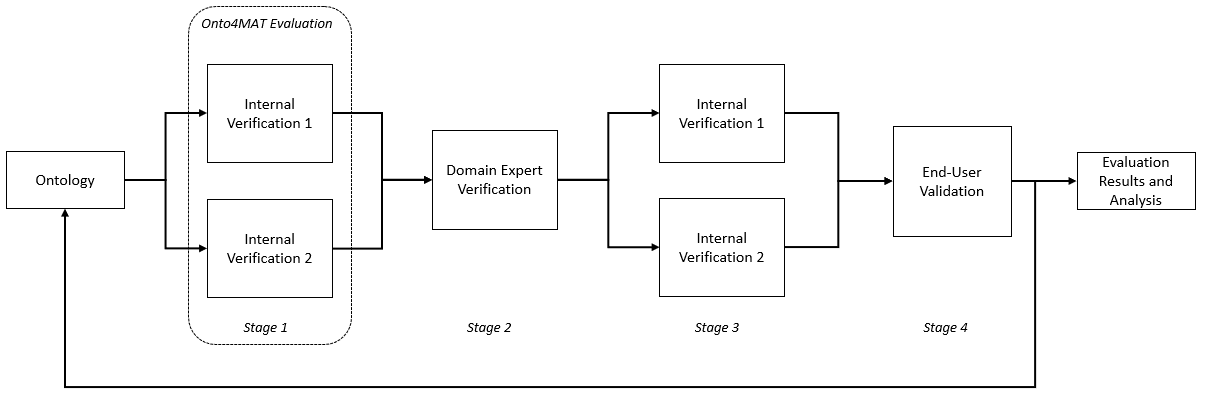}
    \caption{The evaluation methodology developed for this research, highlighting phase 1 (OOPS! and OntoClean) as our evaluation focus in this paper.}
    \label{fig:evalMethod2}
\end{figure*}

Our initial evaluation of Onto4MAT with OOPS! outlined 287 pitfall instances across seven pitfall types, a mix of critical, important and minor problems. Error P4 identified a minor problem with the upper ontology, impacting classes \texttt{Tactic}, \texttt{Agent}, \texttt{Actions}, \texttt{Environment}, and \texttt{AbilitiesAndTraits}. These classes were initially used only for concept organisation as upper ontology elements, without necessary function. This was resolved through the instantiation of object properties, as depicted in Figure~\ref{fig:upperOntology}, that relate these classes to the remainder of the ontology. Error P7 identified a minor problem with our class names, specifically \texttt{BlockAndHold}, \texttt{CastAndMuster}, and \texttt{AbilitiesAndTraits}. OOPS! indicated that these classes may represent multiple concepts that should be represented in their distinct classes. Being domain inspired, Onto4MAT faithfully represents shepherding concepts in an interpretable way. As P7 was a minor error, our solution here was not to separate concepts from the classes but to remove the word \textit{and} from the classes to resolve the issue. Error P8 indicated a minor problem regarding label annotations. While a substantial problem, this was resolved by ensuring all classes possessed an \texttt{rdfs:label} annotation, as well ensuring that class definitions were also included as \texttt{rdfs:comment} annotations.

Three errors related to object properties P11, P13, and P19. Error P11 indicated an important modelling problem in the ontology, where missing domain and range properties were present, impacting eight initial objective properties. Error P13 indicated a minor problem regarding inverse relationships, which were not explicitly defined. Error P19 identified a critical problem where multiple domains and ranges were defined for objective properties, which asserts that all individuals who use the property are instances of all domain and range classes, the intersection of these classes. Identification of these errors resulted in a holistic review of our object properties. We determined that other relations were required to represent the desired richness of relationships and disambiguate the errors. We addressed these issues by developing the final 57 object properties, which possess the richness for bi-directional communication as the intent of Onto4MAT.

Error 41 was the final error indicated by OOPS, identifying an important problem in the ontology. This relates to the absence of an appropriate licence in the metadata, which was resolved with the addition of the Dublin Core (DC) standard for defining rights and licences~\cite{dublincore}.

\begin{table*}[]\label{table:oops}
    \resizebox{\textwidth}{!}{%
    \begin{tabular}{@{}llll@{}}
        \toprule
        \textbf{Error Code} &
          \textbf{Error Description} &
          \textbf{Comment} &
          \textbf{Number of Errors} \\ \midrule
        \multicolumn{1}{c}{P4} &
          \multicolumn{1}{l}{Creating unconnected ontology elements} &
          \multicolumn{1}{l}{Tactic, Agent, Actions, Environment, AbilitiesAndTraits} &
          \multicolumn{1}{c}{5} \\ \midrule
        \multicolumn{1}{c}{P7} &
          \multicolumn{1}{l}{Merging different concepts in the same class} &
          \multicolumn{1}{l}{BlockAndHold, CastAndMuster, AbilitiesAndTraits} &
          \multicolumn{1}{c}{3} \\ \midrule
        \multicolumn{1}{c}{P8} &
          \multicolumn{1}{l}{Missing annotations} &
          \multicolumn{1}{l}{All initial primitive classes} &
          \multicolumn{1}{c}{211} \\ \midrule
        \multicolumn{1}{c}{P11} &
          \multicolumn{1}{l}{Missing domain or range in properties} &
          \multicolumn{1}{l}{testAbilityTo, ensures, inverseOf, oppositeOf, requires, doesA, influencedBy, affectedBy} &
          \multicolumn{1}{c}{8} \\ \midrule
        \multicolumn{1}{c}{P13} &
          \multicolumn{1}{l}{Inverse relationships not explicitly declared} &
          \multicolumn{1}{l}{All initial object properties} &
          \multicolumn{1}{c}{39} \\ \midrule
        \multicolumn{1}{c}{P19} &
          \multicolumn{1}{l}{Defining multiple domains or ranges in properties} &
          \multicolumn{1}{l}{} &
          \multicolumn{1}{c}{20} \\ \midrule
        \multicolumn{1}{c}{P41} &   
          \multicolumn{1}{l}{No license declared} &
          \multicolumn{1}{l}{} &
          \multicolumn{1}{c}{1} \\ \midrule & &
          \multicolumn{1}{l}{} &
          \multicolumn{1}{c}{\textbf{287}} \\ \cmidrule(l){4-4}
    \end{tabular}
    }%
    \caption{Onto4MAT summary statistics.}
\end{table*}

Evaluation of Onto4MAT with OntoClean initially identified 14 meta-property inconsistencies across the ontology, impacting 8\% of ontology classes. These errors were primarily associated with the concept of rigidity, with the majority of classes improperly assigned with the meta-property Non-Rigid (-R). These errors were resolved by assessing and reassigning ABox individuals with alternative meta-properties, which resulted in no further inconsistencies.

\section{Conclusion}\label{sec:conclusion}

In this paper, we have presented a shepherding-based ontology to facilitate multi-agent teaming through formal knowledge representation. Onto4MAT represents biologically-inspired shepherding as a way to share semantic knowledge, promoting bi-directional transparency between humans and artificial agents. Shepherding is an abstraction method for group control where a collection of agents are influenced to achieve the desired intent. Ontologies decompose a target domain with concepts as context to bridge the gap between functional information-sharing systems and semantic understanding, with systemic applications across human-centric functions such as HST. Onto4MAT enables a human to provide an intent as tasks to a collection of agents and for agents to report back to the human. Onto4MAT acts as the epistemological bridge between what is observed and known and for all agents to share a common semantic understanding of the system and environment.

In our subsequent works, we intend to use Onto4MAT for two primary tasks, being the reasoning for action production and recognition. In the first case, the ontology will fulfil the role of a knowledge base to bridge the natural language used by a human user to a collection of artificial agents. The natural language will convey an intent to the shepherding control agent, who will subsequently set forth on a series of self-defined tasks to achieve their interpretation of the desired outcome. This will replace the extant Q\&A approach demonstrated with a broader corpus, accounting for lexical sets. This will be evaluated with a series of domain experts, filling the role of the shepherd (commander) for a broader range of situations and contexts. In recognition tasks, the ontology will be used from the perspective of an external observer to understand the foundations of the system through domain concept identification (contexts), answer questions about entities of the system (situations), and output semantic knowledge for causal discovery (functions). The ontology guides the system's design and organisation of markers with discriminatory power in this setting.

\section*{Acknowledgement}
Thank you to Michael DeBellis for his engaging and thought-provoking discussions on Prot{\'{e}}g{\'{e}} and ontology modelling.

\section*{Data Availability}
Onto4MAT is available on GitHub at \texttt{Fadamos/Onto4MAT}, or http://swarmshepherding.net/software.html

\section*{Declaration of competing interest}
The authors declare that they have no known competing financial interests or personal relationships that could have appeared to influence the work reported in this paper.

\bibliographystyle{elsarticle-num}
\bibliography{bibliography}

\end{document}